\documentclass[lettersize,journal]{IEEEtran}
\usepackage{amsmath,amsfonts}
\usepackage{algorithmic}
\usepackage{array}
\usepackage[caption=false,font=normalsize,labelfont=sf,textfont=sf]{subfig}
\usepackage{textcomp}
\usepackage{stfloats}
\usepackage{url}
\usepackage{verbatim}
\usepackage{graphicx}
\def\BibTeX{{\rm B\kern-.05em{\sc i\kern-.025em b}\kern-.08em
    T\kern-.1667em\lower.7ex\hbox{E}\kern-.125emX}}
\usepackage{balance}
\usepackage{cite}
\usepackage{xcolor}
\usepackage{times}
\usepackage{epsfig}
\usepackage{amsmath}
\usepackage{amssymb}
\usepackage{multirow}
\usepackage{booktabs}
\usepackage[LGRgreek]{mathastext}  
\usepackage{color}
\usepackage{subfig}

\begin{document}
\title{Hyperspectral Image Compression Using\\ Sampling and Implicit Neural Representations}
\author{\IEEEauthorblockN{Shima Rezasoltani and Faisal Z. Qureshi}
\IEEEauthorblockA{Faculty of Science\\
University of Ontario Institute of Technology\\
Oshawa, ON L1H 0C5 Canada\\
\{shima.rezasoltani,faisal.qureshi\}@ontariotechu.ca}}

\maketitle

\begin{abstract}
Hyperspectral images, which record the electromagnetic spectrum for a pixel in the image of a scene, often store hundreds of channels per pixel and contain an order of magnitude more information than a similarly-sized RBG color image. Consequently, concomitant with the decreasing cost of capturing these images, there is a need to develop efficient techniques for storing, transmitting, and analyzing hyperspectral images.  This paper develops a method for hyperspectral image compression using implicit neural representations where a multilayer perceptron network $f_\Theta$ with sinusoidal activation functions ``learns'' to map pixel locations to pixel intensities for a given hyperspectral image $I$.  $f_\Theta$ thus acts as a compressed encoding of this image, and the original image is reconstructed by evaluating $f_\theta$ at each pixel location. We use a sampling method with two factors: window size and sampling rate to reduce the compression time. We have evaluated our method on four benchmarks---Indian Pines, Jasper Ridge, Pavia University, and Cuprite using PSNR and SSIM---and we show that the proposed method achieves better compression than JPEG, JPEG2000, and PCA-DCT at low bitrates. Besides, we compare our results with the learning-based methods like PCA+JPEG2000, FPCA+JPEG2000, 3D DCT, 3D DWT+SVR, and WSRC and show the corresponding results in the "Compression Results" section. We also show that our methods with sampling achieve better speed and performance than our methods without sampling.
\end{abstract}


\section{Introduction}
Unlike a grayscale image that records a single intensity value per pixel, hyperspectral images capture electromagnetic spectrum per pixel~\cite{l1971lightness,goetz1985imaging}.  Therefore, each pixel in a hyperspectral image contains tens or hundreds of values, representing recorded reflectance at various frequency bands.  As a result, hyperspectral images offer greater possibilities for object detection, material identification, and scene analysis than those provided by a typical color RGB image.  The costs associated with capturing high-resolution (both spatial and spectral) hyperspectral images continue to decrease, and it is no surprise that hyperspectral images have found widespread use in areas such as remote sensing, biotechnology, crop analysis, environmental monitoring, food production, medical diagnosis, pharmaceutical industry, mining, and oil \& gas exploration, etc.~\cite{ghamisi2017advances,govender2007review,adam2010multispectral,fischer2006multispectral,liang2012advances,carrasco2003hyperspectral,afromowitz1988multispectral, kuula2012using,schuler2012preliminary,padoan2008quantitative,edelman2012hyperspectral,gowen2007hyperspectral,feng2012application,clark1995mapping}.  Hyperspectral images requires two orders of magnitude more space than what is needed to store a similarly sized color RGB image.  Consequently, there is a need to develop efficient schemes for capturing, storing, transmitting, and analyzing hyperspectral images.  This work studies the problem of hyperspectral image compression, with a view that it serves an important role in the storage and transmission of these images.    

Recently, there has been a surge in interest in learning-based compression schemes.  For example, autoencoders~\cite{Hz94} and rate-distortion autoencoders~\cite{APF18, BLS17} have been used to learn compact representations of the input signals. Here network weights together with the signal signature---latent representation in the case of autoencoders---serve as the compressed representation of the input signal.  Other concurrent works are exploring the use of Implicit Neural Representations (INRs) for signal compression. INRs are particularly well-suited to describe data that lives on an underlying grid, and as such, these offer a new paradigm for signal representation.
In INRs, the goal is to learn a mapping between a location, say an $(x,y)$ pixel location, and the signal value at that location, e.g., the pixel intensity $I[x,y]$.  This mapping is subsequently used to recreate the original signal.  It is as simple as evaluating the INR at various locations.  In the case of INRs, network parameters serve as the learned representation of the input signal.

\begin{figure*}
    \centerline{
        \includegraphics[width=\linewidth]{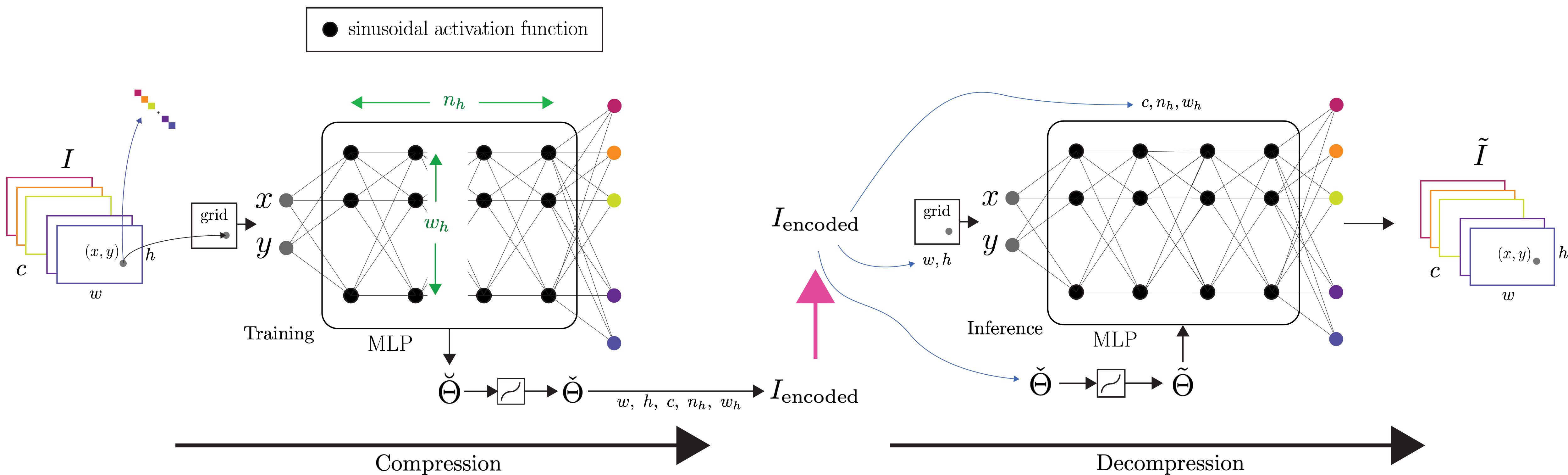}
    }    
   \caption{{\bf Compression and Decompression Pipeline}.  (left) An MLP with a periodic activation function is {\it trained} to map pixel locations to the pixel's spectral signature.  (right) Once fitted, MLP is used to reconstruct the hyperspectral image by performing {\it inference} at various pixel locations.}
    \label{fig:pipeline}
\end{figure*}

\begin{figure}
    \centering
    \includegraphics[width=\linewidth]{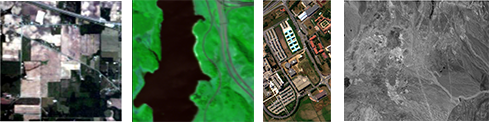}
    \caption{Datasets used in this study shown in pseudo-colors.  (L2R) Indian Pines, Jasper Ridge, Pavia University, and Cuprite.}
    \label{fig:datasets}
\end{figure}

We investigate the use of INRs for hyperspectral image compression and show that it is possible to achieve high rates of compression while maintaining acceptable Peak Signal-to-Noise Ratio (PSNR) values. Figure \ref{fig:pipeline} provides an overview of the proposed compression and decompression pipeline.  We evaluate the proposed approach on four benchmarks (Figure~\ref{fig:datasets})---(1) Indian Pines, (2) Jasper Ridge, (3) Pavia University, and (4) Cuprite---and show that our method achieves better PSNR values than those posted by three popular hyperspectral image compression schemes---(1) JPEG~\cite{good1994joint, qiao2014effective}, (2) JPEG2000~\cite{du2007hyperspectral}, and (3) PCA-DCT~\cite{nian2016pairwise}---at comparable bits-per-pixel-per-band ($bpppb$) values. The results confirm that our method achieves better PSNRs at low compression rates than those obtained by other methods.

The rest of the paper is organized as follows.  We discuss the related work in the next section.  Section~\ref{sec:methodology} describes the proposed method along with evaluation metrics.  Datasets, experimental setup, and compression results are discussed in the following section.   Section~\ref{sec:conclusions} concludes the paper with a summary and possible directions for future work.

\section{Related work}
Hyperspectral images exhibit both spatial and spectral redundancies that can be exploited to achieve compression.  Lossless compression schemes---e.g., those that use quantization or rely upon entropy-coding---where it is possible to recover the original signal precisely often do not yield large savings in terms of memory required to store or transmit a hyperspectral image~\cite{MH12,NV13}.  Lossy compression schemes, on the other hand, promise large savings while maintaining acceptable reconstruction quality.   
Inter-band compression techniques aim to eliminate spectral redundancy~\cite{Cha13}, while intra-band compression techniques aim to exploit spatial correlations.  Intra-band compression techniques often follow the ideas developed for color image compression.  \cite{ZZW16} exploit the fact that groups of pixels that are around the same location in two adjacent bands are strongly correlated and propose schemes that perform both inter-band and intra-band compression.  Principal Component Analysis (PCA) is a popular dimensionality reduction technique that has been used for hyperspectral image compression.  PCA offers strong spectral decorrelation and it can be used to reduce the number of channels in a hyperspectral image. 
The remaining channels are subsequently compressed using Joint Picture Expert Group (JPEG) or JPEG2000 standard~\cite{BS10, DF07, WWJ09, PTM07}.  

Along similar lines, tensor decomposition methods have also been applied to the problem of hyperspectral image compression~\cite{ZZT15}.  Tensor decomposition achieves dimensionality reduction while maintaining the spatial structure.  Transform coding schemes that achieve image compression by reducing spatial correlation have also been used to compress hyperspectral data.  Discrete Cosine Transform (DCT) has been used to perform intra-band compression; however, it ignores inter-band (or spectral) redundancy.  3D-DCT that divides a 3-dimensional hyperspectral image into $8\times8\times8$ datacubes is proposed to achieve both inter-band and intra-band compression~\cite{QRS14}.  Similar to JPEG, which uses $8\times8$ blocks, 3D-DCT exhibits blocking effects in reconstructed hyperspectral images.  The blocking effects can be removed to some extent by using wavelet transform instead~\cite{RSU12,GBS09}.  

Video coding-based methods that treat each channel of a hyperspectral image as a frame in a video have also been used to perform hyperspectral image compression.  These models rely upon inter-band spectral prediction to compress a hyperspectral image.  This is similar to how inter-frame  motion prediction is used for video encoding.

As mentioned earlier, recently, there has been a lot of interest in developing learning-based approaches for signal representation and compression.  E.g., autoencoder-based techniques have been proposed to compress hyperspectral images~\cite{APF18, BLS17}.  Hierarchical variational autoencoders have also been used for the purposes of hyperspectral image compression.  Here the latent variables are discretized for entropy encoding purposes~\cite{balle2018variational, lee2018context, minnen2018joint}.

Work in the area of implicit neural representations has shown that it is possible to represent a signal by overfitting an appropriately designed neural network to it.  Here the parameters (weights) of the neural network serve as the compact representation of the signal, and it is possible to reconstruct the original signal by sampling the neural network at various input locations~\cite{niemeyer2019occupancy, park2019deepsdf, chen2019learning,stanley2007compositional,mildenhall2020nerf,sitzmann2020implicit,tancik2020fourier}.  \cite{dupont2021coin} shows that implicit neural representations with periodic activation functions are able to represent signals, including images, with high-fidelity.  This work serves as an inspiration for our work.

Similar to previous research on latent variable models \cite{hjelm2016iterative, kim2018semi, krishnan2018challenges, marino2018iterative}, numerous studies \cite{campos2019content, guo2020variable, yang2020improving} make an effort to close the amortization gap \cite{cremer2018inference} by combining the usage of amortized inference networks with iterative gradient-based optimization procedures. Using inference time per instance optimization, \cite{yang2020improving} also identifies and makes an effort to bridge the discretization gap caused by quantizing the latent variables. The concept of per-instance model optimization is expanded upon in \cite{van2021overfitting}, which fine-tunes the decoder for each instance and transmits updates to the quantized decoder's parameters together with the latent code to provide better rate-distortion performance.

\section{Image Compression using INRs}
\label{sec:methodology}

Let us consider a $w$-by-$h$ grayscale image.  We can represent this image as a function
$$
I_\text{grayscale}:U \mapsto [0,1],
$$
where 
$$
U=\{(1,1),\cdots,(w,1),\cdots,(1,h),\cdots,(w,h)\}.
$$
This notation captures the intuition that an image is a function over a 2d grid that defines the pixel locations.  The intensity at each pixel is then $I_\text{grayscale}(x,y)$.  It is straightforward to extend this notation to hyperspectral image $I$ as follows
\begin{equation}
I =
\begin{pmatrix}
I_1:U \mapsto [0,1] \\
\vdots\\
I_c:U \mapsto [0,1]
\end{pmatrix}.
\label{eq:hsi-representation}
\end{equation}
Here for the sake of simplicity, we assume a $w$-by-$h$ hyperspectral image comprising $c$ channels.  Using this notation, we can find the spectral signature of the pixel at location $(x,y) \in U$ as follows:
$$
\{I_1(x,y), \cdots, I_c(x,y) \}.
$$
Given this setup, it is possible to imagine a neural network that models the functions $I_1,\cdots,I_c$.  Specifically, work on implicit neural representations suggests using a multilayer perceptron (MLP) with periodic activation functions to represent functions of the form shown in Equation~\ref{eq:hsi-representation}.  Consider an MLP $f_\Theta$ with parameters $\Theta$ that maps locations in $U$ to pixel spectral signatures:  
$$
f_\Theta: U \mapsto \{ I_1,\cdots,I_c \}.
$$
Under this regime, training can be defined as
$$
\breve{\Theta} = \arg\min_\Theta \mathcal{L}(I, f_\Theta),
$$
Where $\mathcal{L}$ is a loss function that is differentiable and that captures the error between the original hyperspectral image and the decompressed hyperspectral image.  We use the mean-squared error to compute this loss.  Given $\breve{\Theta}$, it is possible to reconstruct the original image $I$ by sampling $f_{\breve{\Theta}}$ at the relevant locations.  Parameters $\breve{\Theta}$, along with $w$, $h$, and $c$, plus the structure of the MLP, represent an encoding $I_{\text{encoded}}$ of the hyperspectral image $I$ that was used to train the MLP $f_\Theta$.  It is expected that the memory required to store $I_{\text{encoded}}$ is an order of magnitude less than the memory needed to store the hyperspectral image.  
The decompression process requires constructing the sampling locations $U$, setting up MLP $f_\Theta$ and initializing its weights to $\breve{\Theta}$, and evaluating $f_{\breve{\Theta}}$ at locations in $U$.

\begin{table}
\centering
\begin{tabular}{|c|c|c|c|c|c|c|c|} \hline
    & $w$ & $h$ & $c$ & $n_h$ & $w_h$ & $q$ & $\Theta$ \\\hline
    \#bits & 16 & 16 & 16 & 8 & 8 & 1 & $bpp \times n_\Theta$  \\\hline
\end{tabular}
\caption{Disk layout for $I_{\text{encoded}}$.  Here $q$ denotes if parameters $\Theta$ were quantized at compression time.  $bpp$ (or \#bits-per-parameter) is either $32$ or $16$.}
\label{tbl:encoding}
\end{table}

\begin{figure*}[!t]
    \centering
    \subfloat[Indian Pines]{\includegraphics[width=0.25\textwidth]{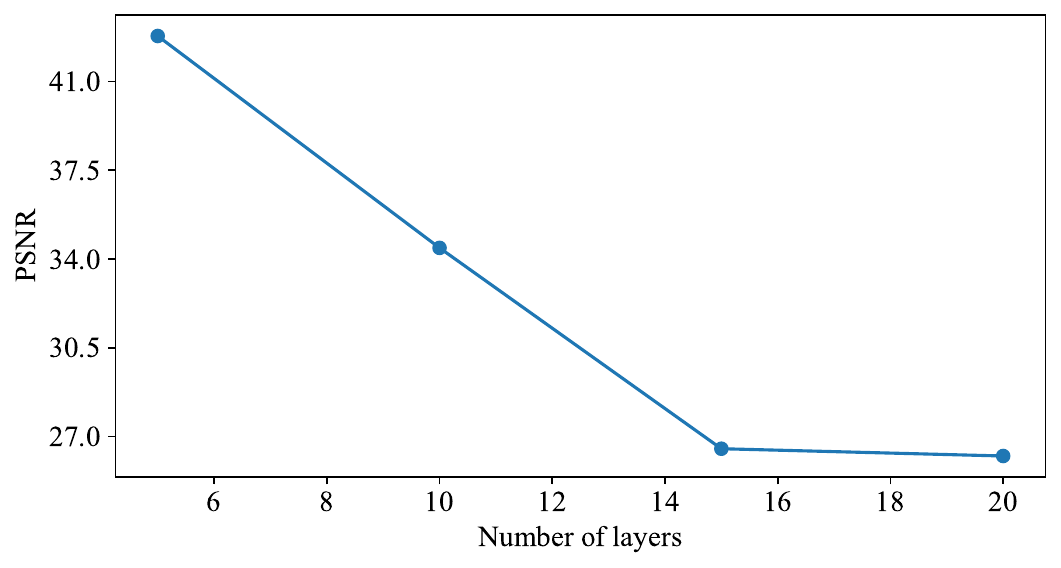}%
    \label{fig_first_case}}
    \hfil
    \subfloat[Jasper Ridge]{\includegraphics[width=0.25\textwidth]{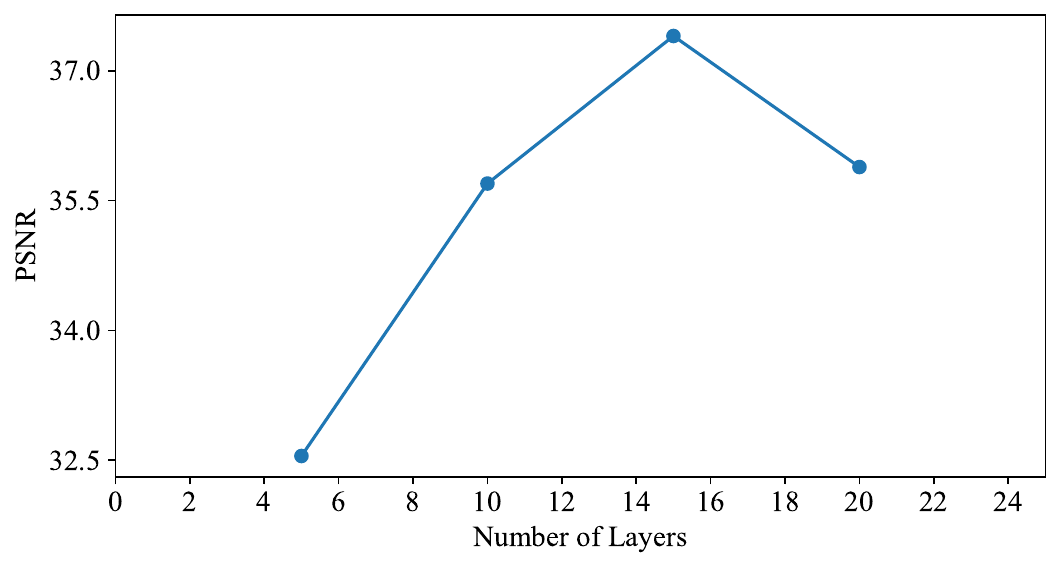}%
    \label{fig_second_case}}
    \hfil
    \subfloat[Pavia University]{\includegraphics[width=0.25\textwidth]{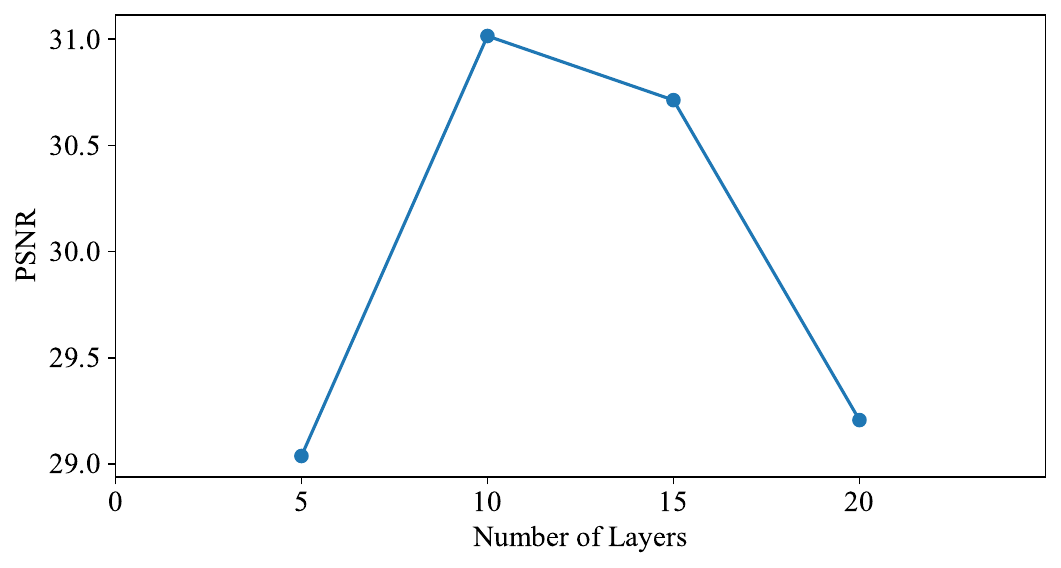}%
    \label{fig_second_case}}
    \hfil
    \subfloat[Cuprite]{\includegraphics[width=0.25\textwidth]{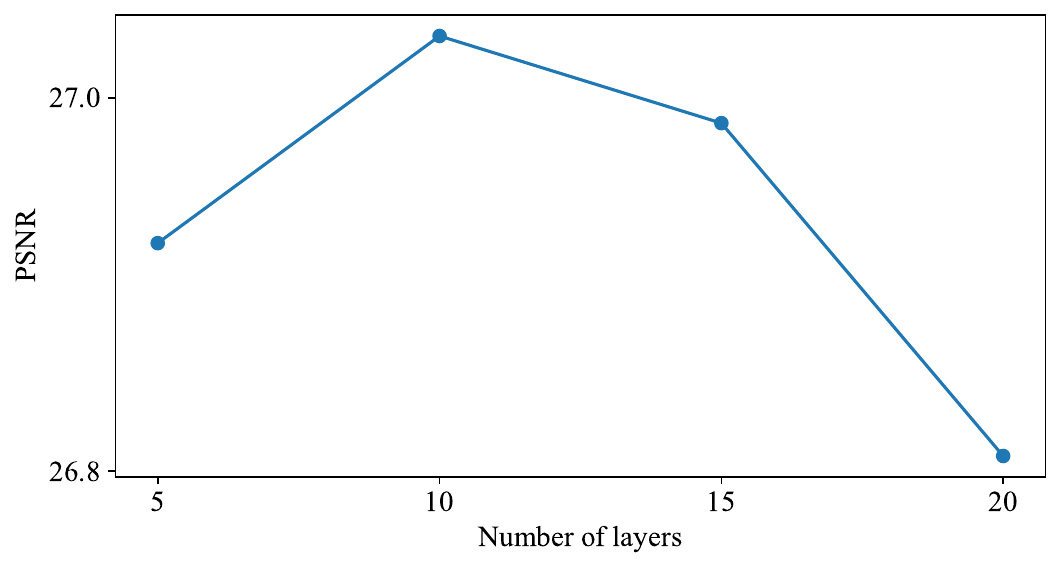}%
    \label{fig_second_case}}
    \caption{{\bf Architecture search}.  Exploring MLP structure that achieves the best PSNR for different datasets (for a fixed $bpppb$ budget).  For our purposes, the MLP structure is defined by the number of hidden layers and the width of these layers.  Together the number and width of the hidden layers define network capacity.}
    \label{fig:model-search}
\end{figure*}

\subsection{Metrics for Measuring Compression Quality}

Similar to previous studies, we use Peak Signal-to-Noise Ratio (PSNR) to compare the proposed method with JPEG, JPEG2000, and PCA-DCT approaches for hyperspectral image compression.  PSNR, measured in decibels, is a frequently used metric in image compression.  It measures the difference in ``quality'' between the original image and its compressed version.  Higher PSNR values suggest that the compressed image is more similar to the original image, i.e., the compressed image preserves more of the information present in the original image and that it has higher quality.  In addition, we also compare the compressed image using Mean Squared Error (MSE), which computes the cumulative error between the original image and its compressed version.  Lower values of MSE mean better reconstruction quality. 

MSE is computed as follows
\begin{equation}
    \label{eq:mse}
    MSE = \sum_i \frac{|I[i]-\tilde{I}[i]|^2}{i},
\end{equation}
where $\tilde{I}$ denotes the compressed image and $i$ indices over the pixels.  MSE is used to calculate PSNR
\begin{equation}
    \label{eq:psnr}
    PSNR = 10\ \log_{10}\left(\frac{R^2}{MSE}\right),
\end{equation}
where $R$ is the largest variation in the input image in the previous equation. For instance, $R$ is $1$ if the input image is of the double-precision floating-point data. $R$ is $255$, for instance, if the data is an 8-bit unsigned integer.

Moreover, We calculate the structure similarity (SSIM) to compare the proposed method with other methods. SSIM is a much newer equation developed in 2004 \cite{wang2004image}. SSIM is based on the computation of three factors; luminance, contrast, and structure. The overall index is a multiplicative combination of the three. The SSIM is a mean value obtained by averaging the SSIM values of all bands and is used for the evaluation of spatial structure preservation. The PSNR measures the proximity of the original image to its approximation, and the SSIM measures the visual quality of the approximation image. Note that larger SSIM values mean better reconstruction performance. The formula for calculating SSIM is given in the below Equation,

\begin{equation}
    \label{eq:ssim}
    SSIM(x,y) = \frac{(2\ \mu_{x}\ \mu_{y}\ +\ C_{1})\ (2\ \sigma_{xy}\ +\ C_{2})}{(\mu_{x}^2\ +\ \mu_{y}^2\ +\ C_{1})(\sigma_{x}^2\ +\ \sigma_{y}^2\ +\ C_{2})},
\end{equation}
where $\mu_{x}$ and $\mu_{y}$ are estimated as the mean of each image $x$ and $y$. $\sigma_{x}^2$, $\sigma_{y}^2$ are the variance and $\sigma_{xy}$ is the covariance. $C_{1} = (k_{1}.L)^2$ and $C_{2} = (k_{2}.L)^2$ are variables to stabilize the division with a weak denominator. $L$ is the dynamic range of the pixel-values (typically, this is $2^{bitsperpixel}$ and by default, $K_{1} = 0.01$ and $K_{2} = 0.03$.
The SSIM values range from 0 to 1, where 1 is a perfect match between the original and reconstructed images.

In addition, the number of bits-per-pixel-per-band ($bpppb$) captures the level of compression achieved by a model.  Lower values of $bpppb$ indicate higher compression rates.  The $bpppb$ value for an uncompressed hyperspectral image is either $8$ or $32$ bits, depending upon how the pixels are stored.  It is common to store hyperspectral pixels value (for each channel) as a $32$-bit floating point.  The parameter $bpppb$ is calculated as follows:
\begin{equation}
    \label{eq:bpppb}
    bpppb = \frac{\#\mathrm{parameters} \times (\mathrm{bits\ per\ parameter})}{(\mathrm{pixels\ per\ band}) \times \#\mathrm{bands}}.
\end{equation}

\begin{figure*}[!t]
    \centering
    \subfloat[Indian Pines]{\includegraphics[width=0.25\textwidth]{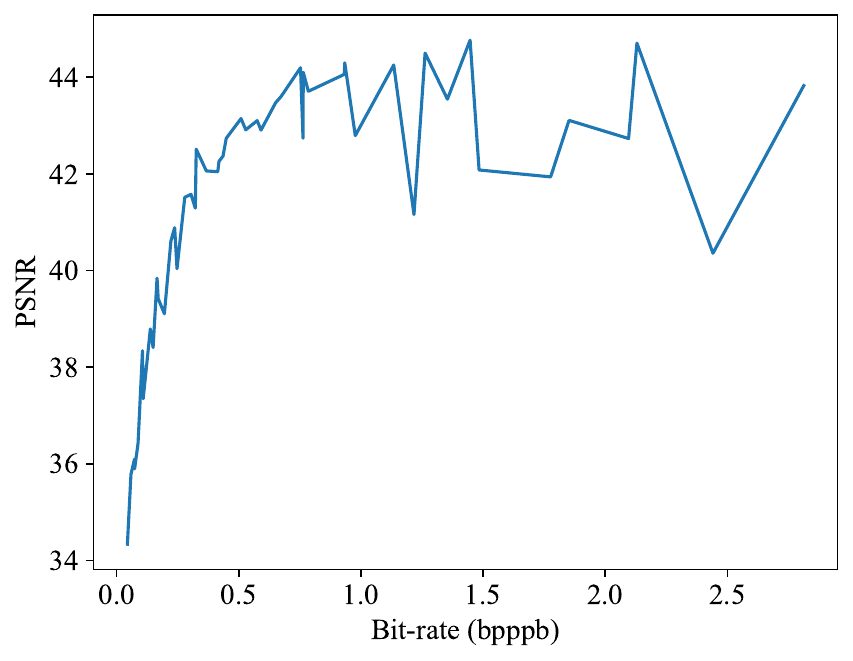}%
    \label{fig_first_case}}
    \hfil
    \subfloat[Jasper Ridge]{\includegraphics[width=0.25\textwidth]{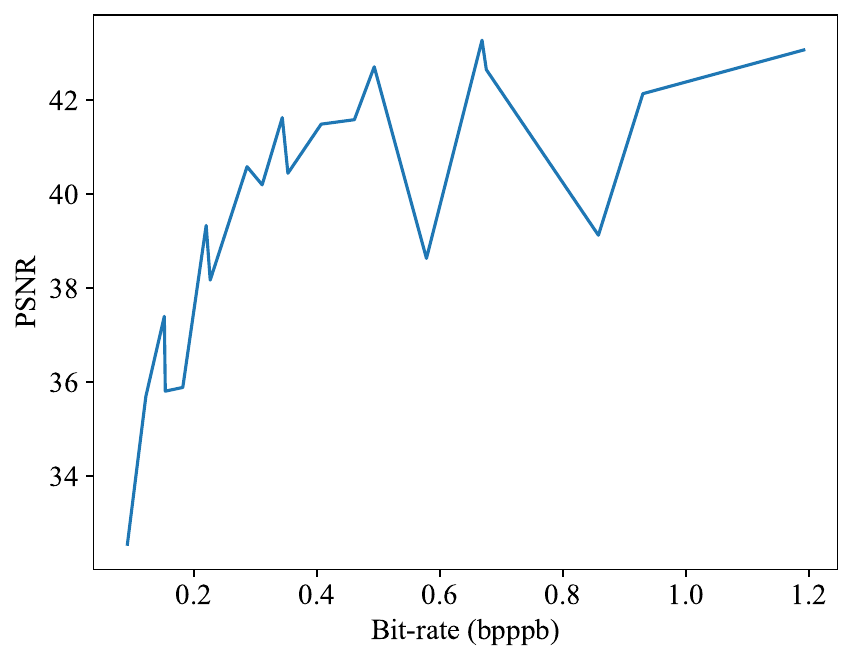}%
    \label{fig_second_case}}
    \hfil
    \subfloat[Pavia University]{\includegraphics[width=0.25\textwidth]{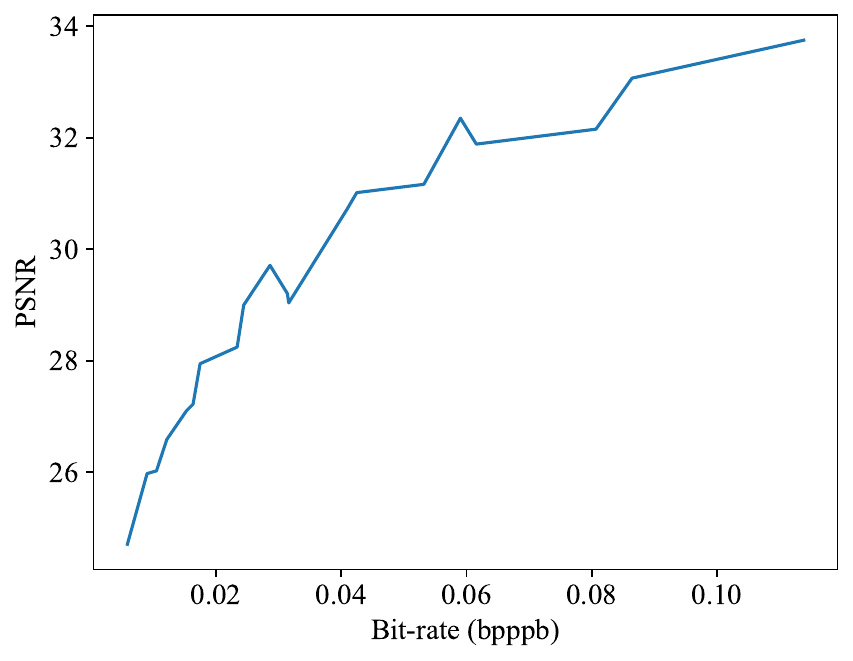}%
    \label{fig_second_case}}
    \hfil
    \subfloat[Cuprite]{\includegraphics[width=0.25\textwidth]{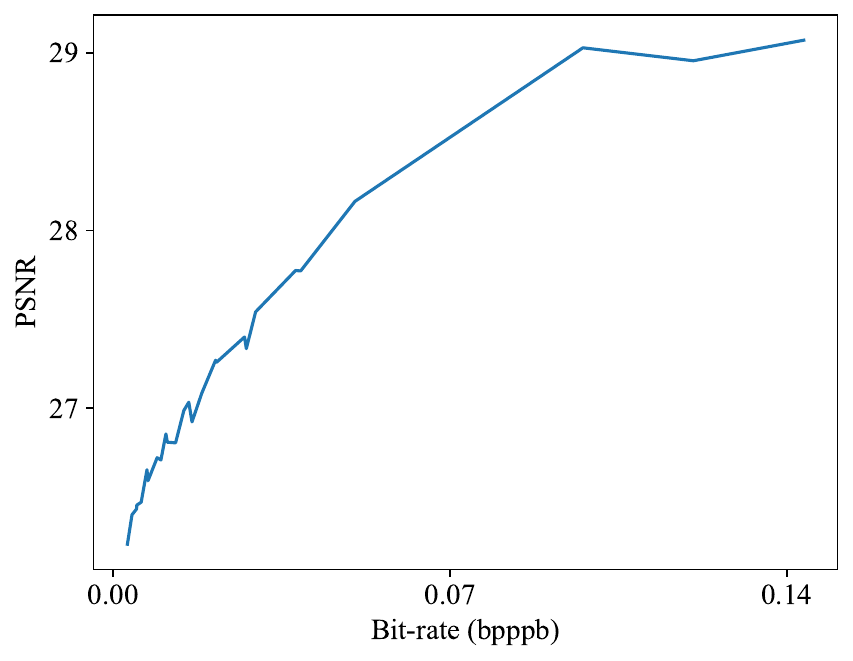}%
    \label{fig_second_case}}
    \caption{{\bf Model capacity}. PSNR vs. $bpppb$. The trend of these plots confirms our intuition that PSNR values increase as $bpppb$ numbers are increased.  The plots are not monotonically non-decreasing.  This has to do with the stochastic nature of MLP overfitting.}
    \label{fig:psnr-vs-bpppb}
\end{figure*}

Figure~\ref{fig:psnr-vs-bpppb} plots PSNR vs. bpppb for the four datasets that we are using in this work.  The plots confirm our intuition that higher $bpppb$ leads to better compression quality as measured by PSNR values.  For our method, $bpppb$ calculations do not include the storage required to keep network structures.

\subsection{Compression Pipeline}

The proposed compression method consists of two steps.  Step 1 performs an architecture search.  The goal here is to find an MLP that achieves the highest reconstruction accuracy for a given $bpppb$ budget.  Architecture search is performed by overfitting multiple MLPs having different numbers of hidden layers and hidden layers' widths to the hyperspectral image.  Architecture search, however, means longer compression times.  
Step 2 involves quantizing and storing the parameters of the overfitted MLP to disk.  The caveat here is that this further reduces the quality of the reconstructed image.

\subsubsection{Overfitting a SIREN network}

The compression procedure comprises overfitting a SIREN network $f_\Theta$ to a hyperspectral image $I$~\cite{dupont2021coin}.  The width $w$ and height $h$ of the hyperspectral image are used to set up an input location grid on $[-1,+1] \times [-1,+1]$, and the MLP is trained to reconstruct a pixel's spectral signature given its location.
The parameters $\breve{\Theta}$ of this overfitted MLP are quantized $\check{\Theta}$.  MLP structure that contains the number of hidden layers $n_h$, widths of these layers $w_h$, and the width $w$, height $h$, and the number of channels $c$ of the original hyperspectral image $I$ along with $\check{\Theta}$ serve as a compressed encoding $I_{\text{encoded}}$ of the hyperspectral image $I$ (Table~\ref{tbl:encoding}).  Parameters $\check{\Theta}$ are either stored as $32$-bit floats or as $16$-bit floats.  Training and inference require $32$-bit floats, and quantization/dequantization is performed to move between $32$ and $16$ bits representations.  We have yet to try an 8-bit, fixed-point representation for parameters.  

The overfitted MLP contains 
\begin{equation}
(w_h \times 2) + (w_h \times w_h)^{(n_h - 1)} + (c \times w_h)
\label{eq:number-of-parameters}
\end{equation}
parameters.

\subsubsection{Decompressing $I_{\text{encoded}}$}

The hyperspectral image is reconstructed from its compressed encoding $I_{\text{encoded}}$ as follows: 1) use $n_h$, $w_h$, and $c$ to reconstruct $f_\Theta$, 2) dequantize $\check{Q}$  to $\tilde{\Theta}$ and use it to initialize the parameters of $f_\Theta$, 3) use the width $w$ and height $h$ to set up the input grid between $[-1,+1] \times [-1,+1]$, and 4) evaluate $f_{\tilde{\Theta}}$ at each location in the input grid to reconstruct the image $\tilde{I}$.

\begin{figure*}[!t]
    \centering
    \subfloat[Indian Pines]{\includegraphics[width=0.50\textwidth]{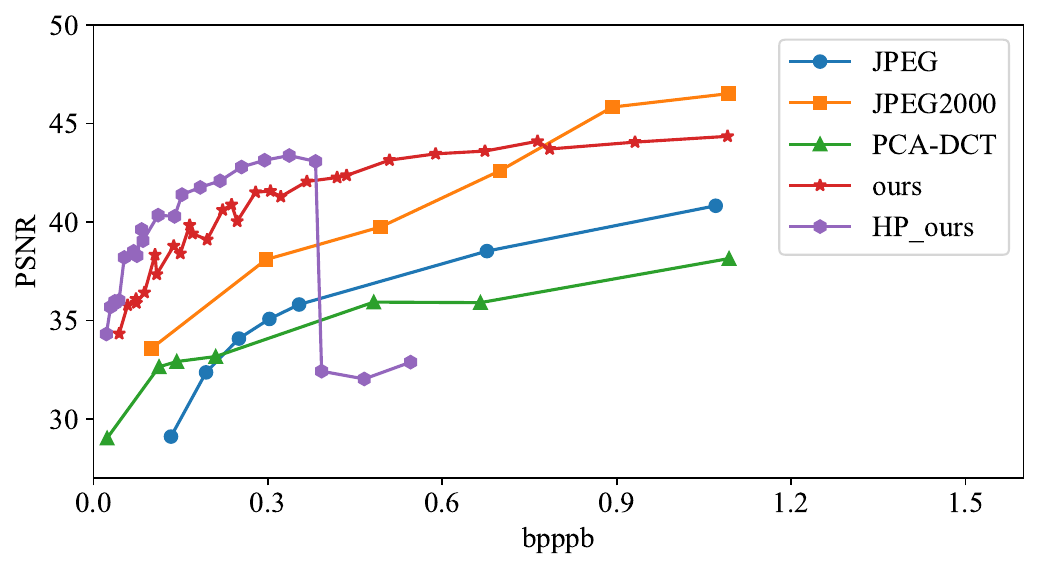}%
    \label{fig_first_case}}
    \hfil
    \subfloat[Jasper Ridge]{\includegraphics[width=0.50\textwidth]{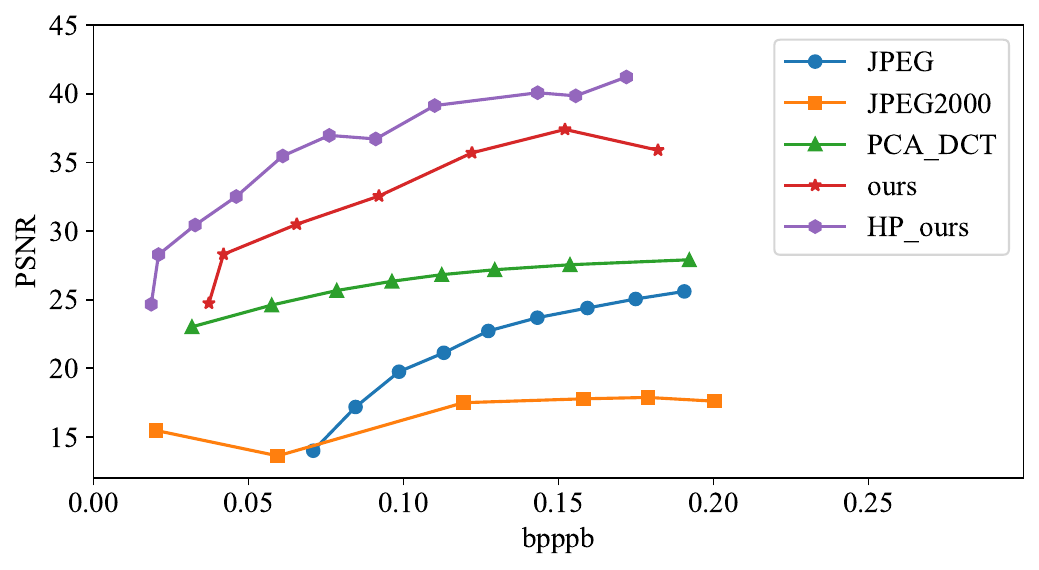}%
    \label{fig_second_case}}
    \hfil
    \subfloat[Pavia University]{\includegraphics[width=0.50\textwidth]{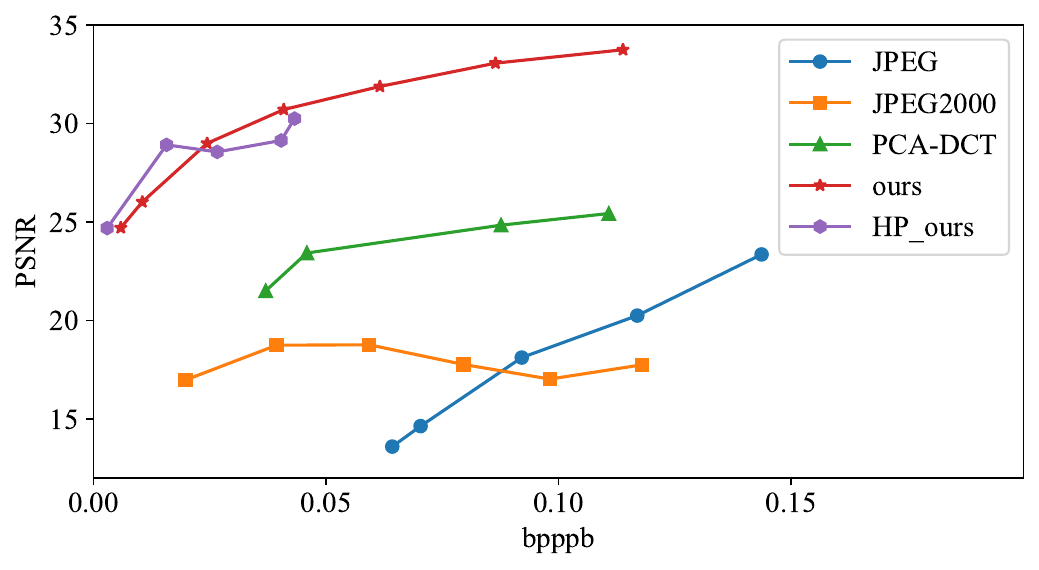}%
    \label{fig_second_case}}
    \hfil
    \subfloat[Cuprite]{\includegraphics[width=0.50\textwidth]{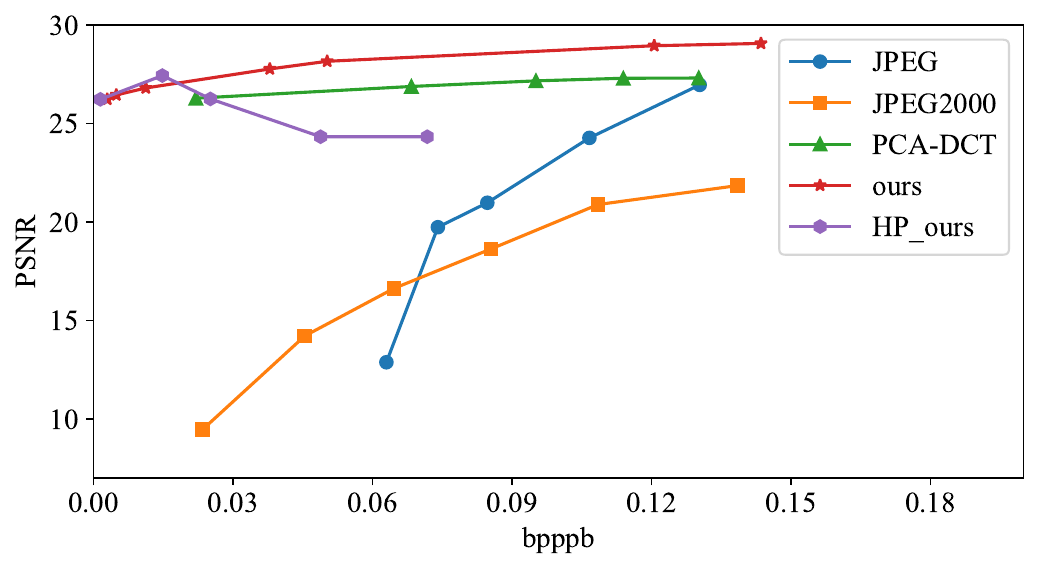}%
    \label{fig_second_case}}
    \caption{{\bf Compression results}.  PSNR values achieved at various $bpppb$ for our method, along with those obtained by JPEG, JPEG2000, and PCA-DCT schemes.  Here ``ours'' refer to our method where parameters are stored at $32$-bit precision, and ``HP\_ours'' refer to results when parameters are stored at $16$-bit precision.}
    \label{fig:psnr-vs-bpppb-comparison}
\end{figure*}

\section{Experiments}
\label{sec:experiments}

We have used four datasets to evaluate our approach: 

\begin{itemize}

\item {\bf Indian Pines:}
This is a $145 \times 145 \times 220$ hyperspectral image.  It was collected using the AVIRIS sensor in 1992, and it spans a region over NW Indiana.  The hyperspectral image contains a mix of farms and wooded areas.  Additionally, the image contains low-density built-up regions, houses, a number of secondary roads, a rail line, and two dual-lane motorways.  

\item {\bf Jasper Ridge:}
This is a $100 \times 100 \times 224$ hyperspectral image.  It was also captured using the AVIRIS sensor.

\item{\bf Pavia University:}
It is a $610 \times 340 \times 103$ hyperspectral image.  This dataset was captured by the ROSIS-03 aerial instrument that was flown by the German Aerospace Centre as part of the HySens project.  

\item {\bf Cuprite:}
The Cuprite dataset contains one $614 \times 512 \times 224$ hyperspectral image.

\end{itemize}

We have selected these datasets since others have used these previously to study hyperspectral image compression.  These datasets have been collected using NASA's Airborne Visible/Infrared Imaging Spectrometer (AVIRIS) sensor.  The AVIRIS sensor gathers geometrically coherent spectro-radiometric data that can be used to characterize the Earth's surface.  The captured data has found use in areas ranging from oceanography, snow hydrology, geology and volcanology, and limnology to environmental studies, atmospheric and aerosol studies, agriculture, and land management and use.  

\subsection{Setup}

We compare our work with three hyperspectral image compression methods: 1) JPEG ~\cite{good1994joint, qiao2014effective}; 2) PCA-DCT~\cite{nian2016pairwise}, and 3) JPEG2000~\cite{du2007hyperspectral}.  JPEG method for hyperspectral image compression uses JPEG standard to encode each channel (band) separately.  
JPEG2000, instead, uses the JPEG2000 standard for encoding the hyperspectral image.  It, too, treats each channel separately.  PCA-DCT uses PCA-based analysis to reduce the number of channels, followed by a DCT-based method for encoding these channels.  PCA-DCT method posts low signal-to-noise ratios; however, this can be fixed somewhat by keeping more of the original channels.  We have chosen these hyperspectral image compression techniques since they are widely used for reducing the size of hyperspectral data in hyperspectral analysis pipelines. Besides, we compare our results with the learning-based methods like PCA+JPEG2000, FPCA+JPEG2000, 3D DCT, 3D DWT+SVR and WSRC, and show the corresponding results in the "Compression Results" section.

\subsection{Architecture Search}

Given an image and our (MLP) parameter budget, which is measured in bits per pixel per band, or $bpppb$ for short), the first goal is to select the MLP structure---i.e., the number of hidden layers and their widths---that is able to represent this image with an acceptable PSNR value.  Figure~\ref{fig:model-search} shows PSNR values achieved for different architectures for the four datasets having a fixed $bpppb$ budget.  This suggests that network structure, in addition to network capacity, affects how well a network represents the hyperspectral image. 

MLP structure is chosen via hyperparameter search, which involves training feasible designs containing the right number of hidden layers having the correct width on a given hyperspectral image.  The result of this process is a single MLP that is able to reconstruct the hyperspectral image with the desired PSNR value.  The parameters of the final MLP are then quantized to $16$-bit precision, which leads to further savings in terms of the storage needed to represent the hyperspectral image.  Our experiments suggest that reducing the MLP parameters from $32$-bit to $16$-bit precision did not increase distortion and that it had little effect on the signal-to-noise ratio.

\begin{figure*}[!t]
    \centering
    \subfloat[Model training on Indian Pines dataset at 0.2 $bpppb$.]{\includegraphics[width=0.25\textwidth]{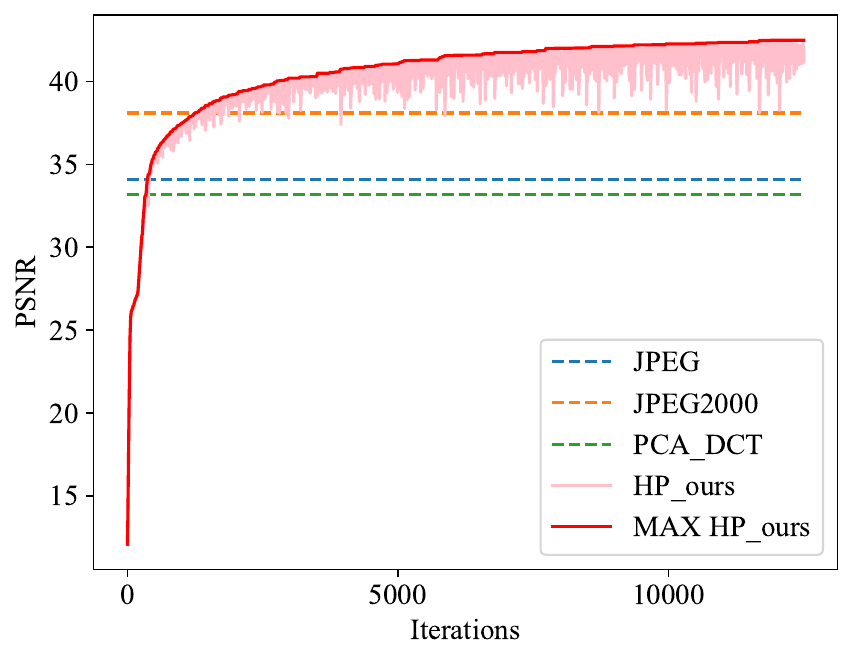}%
    \label{fig_first_case}}
    \hfil
    \subfloat[Model training on Jasper Ridge dataset at 0.15 $bpppb$.]{\includegraphics[width=0.25\textwidth]{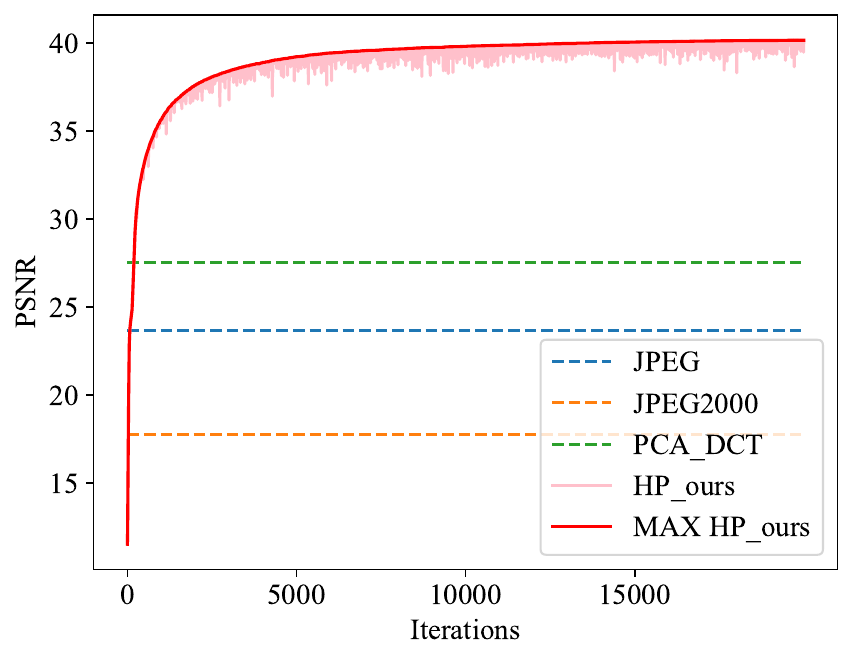}%
    \label{fig_second_case}}
    \hfil
    \subfloat[Model training on Pavia University dataset at 0.025 $bpppb$.]{\includegraphics[width=0.25\textwidth]{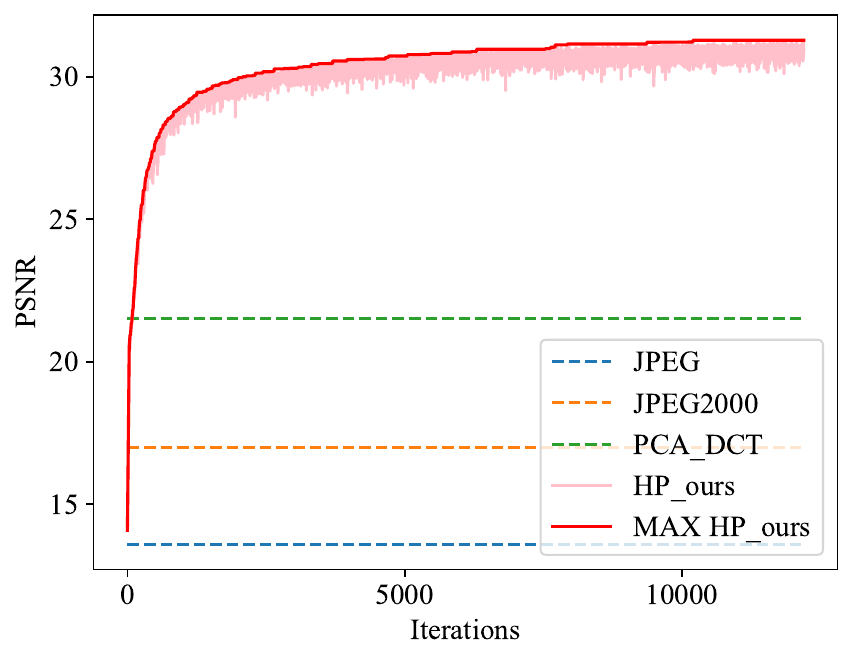}%
    \label{fig_second_case}}
    \hfil
    \subfloat[Model training on Cuprite dataset at 0.02 $bpppb$.]{\includegraphics[width=0.25\textwidth]{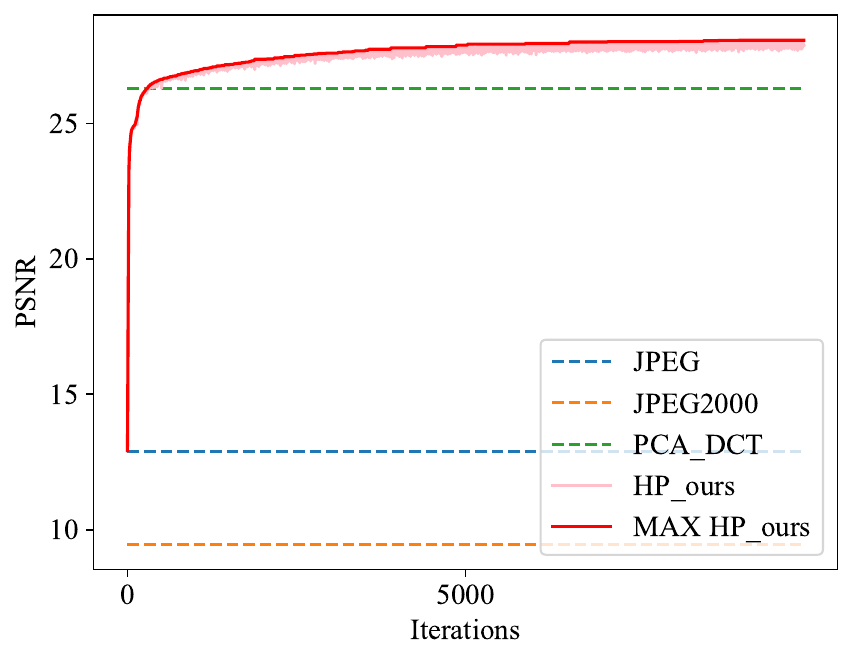}%
    \label{fig_second_case}}
    \caption{Encoding procedure.  Model training on (counter-clockwise from top-left) Indian Pines, Jasper Ridge, Pavia University, and Cuprite datasets.  At around $2000$ iteration mark, our method is already achieving better PSNR values than those for JPEG, JPEG2000, and PCA-DCT.  Furthermore, the PSNR value for our methods continues to improve with more iterations (up to a point).}
    \label{fig:overfitting}
\end{figure*}

\subsection{Comparison with other methods}

Figure~\ref{fig:psnr-vs-bpppb-comparison} shows PSNR values at various compression rates for different methods.  Specifically, we compare our approach, labeled as \emph{ours} and \emph{HP\_ours}, with JPEG, JPEG2000, and PCA-DCT methods.  Here, \emph{ours} method stores MLP weights as $32$-bit floating point values, whereas \emph{HP\_ours} stores MLP weights at half-precision as $16$-bit floating point values that are constructed by quantizing the MLP weights.  These plots illustrate that our methods achieve higher compression quality, i.e., better PSNR, for a given value of $bpppb$.  This is especially true for lower $bpppb$ values.

For the Indian Pines dataset, \emph{ours} method achieves better PSNR up to around $0.7$ $bpppb$, at which point JPEG2000 obtains better PSNR.  What is curious is that the PSNR for\emph{HP\_ours} drops drastically at around $0.4$ $bppb$. This merits further investigation.  For Jasper Ridge, \emph{HP\_ours} performs better than \emph{ours}.  However, both \emph{ours} and \emph{HP\_ours} achieve higher PSNR values than other methods.  For Pavia University and Cuprite datasets, \emph{our} method obtains better PSNR values than other methods.

We draw the following conclusions from these results: 1) the proposed method, both \emph{ours} and \emph{HP\_ours}, perform high-quality compression at high compression rates; 2) it is beneficial to perform architecture search plus examine the effects of quantization at compression time since on some datasets \emph{HP\_ours} outperforms \emph{ours}; and 3) the compression quality obtained by the proposed method compares favorably with the three commonly used compression methods for hyperspectral images.

\subsection{Encoding Considerations}

Our method belongs to the class of ``slow-encoding-fast-decoding'' compression methods.  The method needs to train, actually \emph{overfit}, multiple MLPs at encoding (compression) time.  This is needed to find the MLP structure that best represents the hyperspectral image given a particular storage budget.  Decoding, however, only requires evaluating this MLP at various pixel locations.  Decoding is fast.  It can be made even faster by exploiting the parallelism inherent to this procedure.  The ``slow-encoding-fast-decoding'' nature of this method makes it particularly suitable for applications where the hyperspectral image is compressed once only, say at capture time.  

We show an example of the overfitting procedure in Figure~\ref{fig:overfitting}. These plots show the encoding procedure on the four datasets: (1) Indian Pines at $0.2$ $bpppb$; (2) Jasper Ridge at $0.15$ $bpppb$; (3) Pavia University at $0.025$ $bpppb$; and (4) Cuprite at $0.02$ $bpppb$.  JPEG, JPEG2000, and PCA-DCT methods do not require iterations. Consequently, their respective PSNR values are denoted with the horizontal dashed lines.  The method proposed in this paper is iterative.  Note that PSNR values for \emph{HP\_ours} continue to increase with the number of iterations (up to a point).   Improvement in PSNR values saturates at around $10,000$, $15,000$, $10,000$, and $5,000$ iterations for Indian Pines, Jasper Ridge, Pavia University, and Cuprite datasets, respectively.  This hints at the upper bound on encoding, or compression, time for our method.  Note also that at around $2,000$ iteration mark \emph{HP\_ours} method starts to obtain better PSNR values than the other three methods.  As stated earlier, our method involves model fitting, which is inherently stochastic.   Therefore, throughout the iterative process, we store the model parameters that obtained the  highest value for PSNR thus far.  In these plots, \emph{MAX HP\_ours} denote these PSNR scores. This guarantees that the model does not get worse over time.    

\subsection{Model Fitting}

The number of inputs for all our models was $2$, and the number of outputs was equal to the number of channels (or bands) of the hyperspectral image.  The activation functions for hidden layers were sinusoidal.  We initialized the MLP using the guidelines provided in~\cite{sitzmann2020implicit}.  Adam optimizer was used during training, and the learning rate was set to $2e-4$.  All experiments were conducted on an Intel i7 desktop with Nvidia RTX 2080 GPU.

\begin{figure*}[!t]
    \centering
    \subfloat{\includegraphics[width=0.11\textwidth]{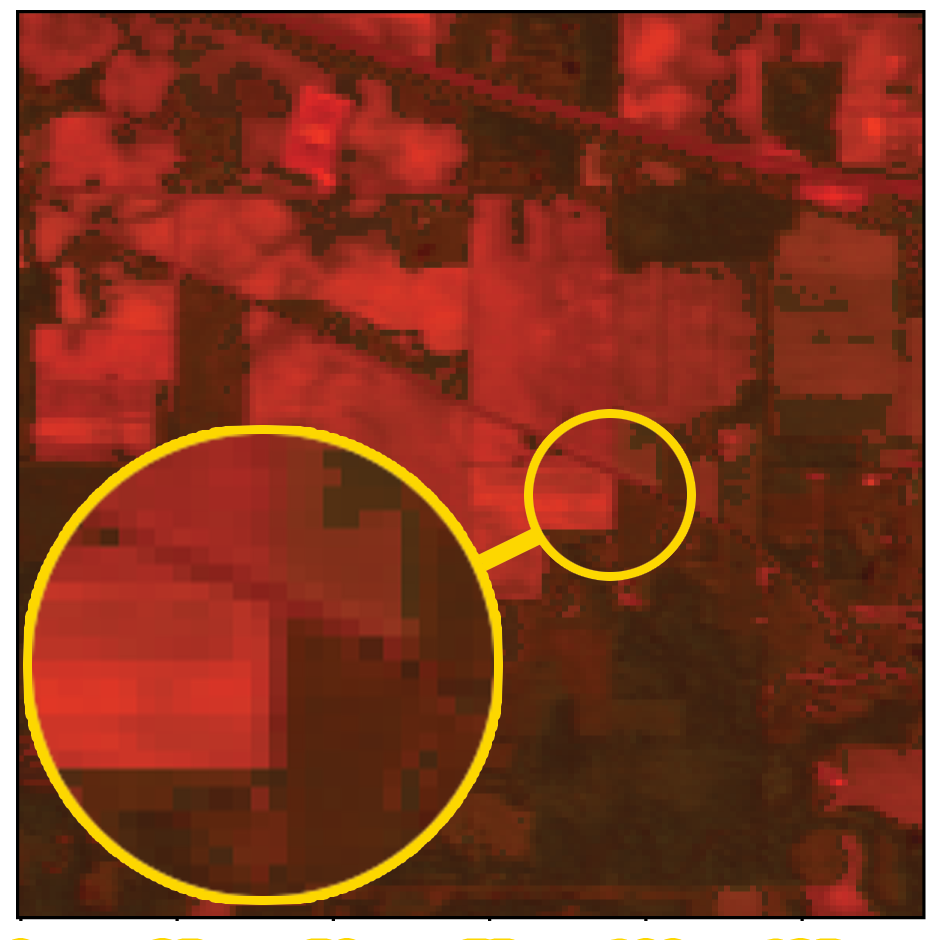}%
    \label{fig_first_case}}
    \subfloat{\includegraphics[width=0.11\textwidth]{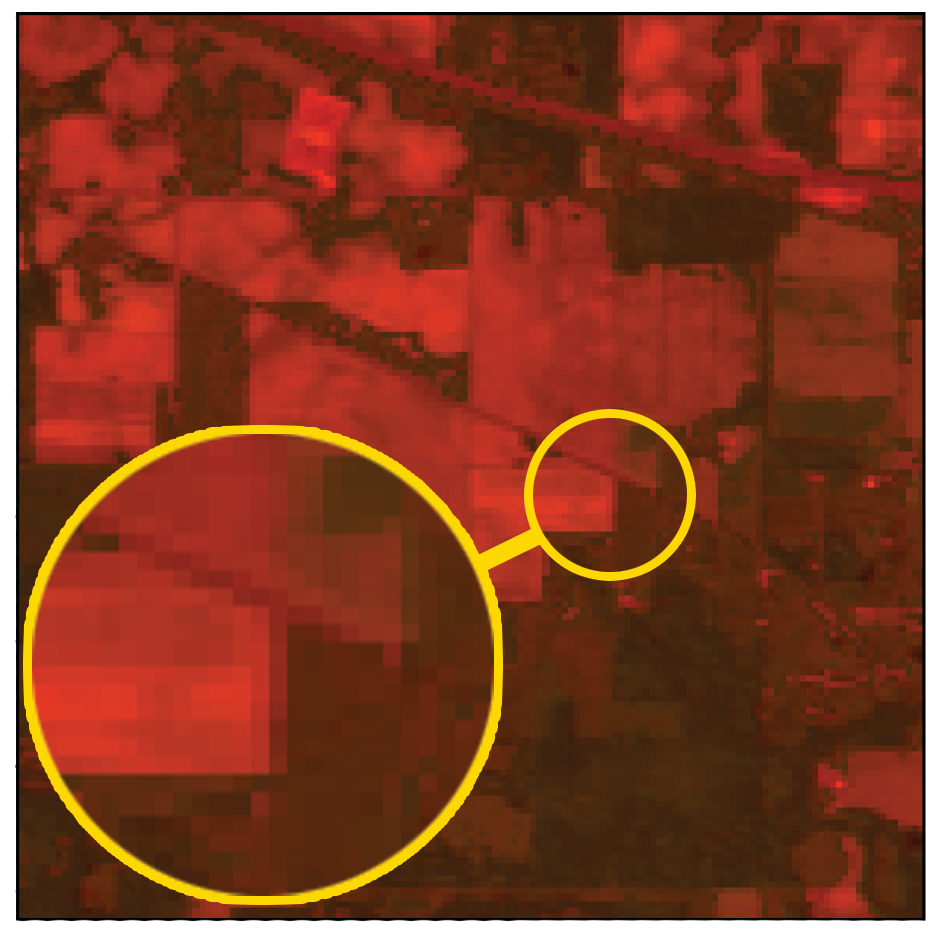}%
    \label{fig_first_case}}
    \hfil
    \subfloat{\includegraphics[width=0.11\textwidth]{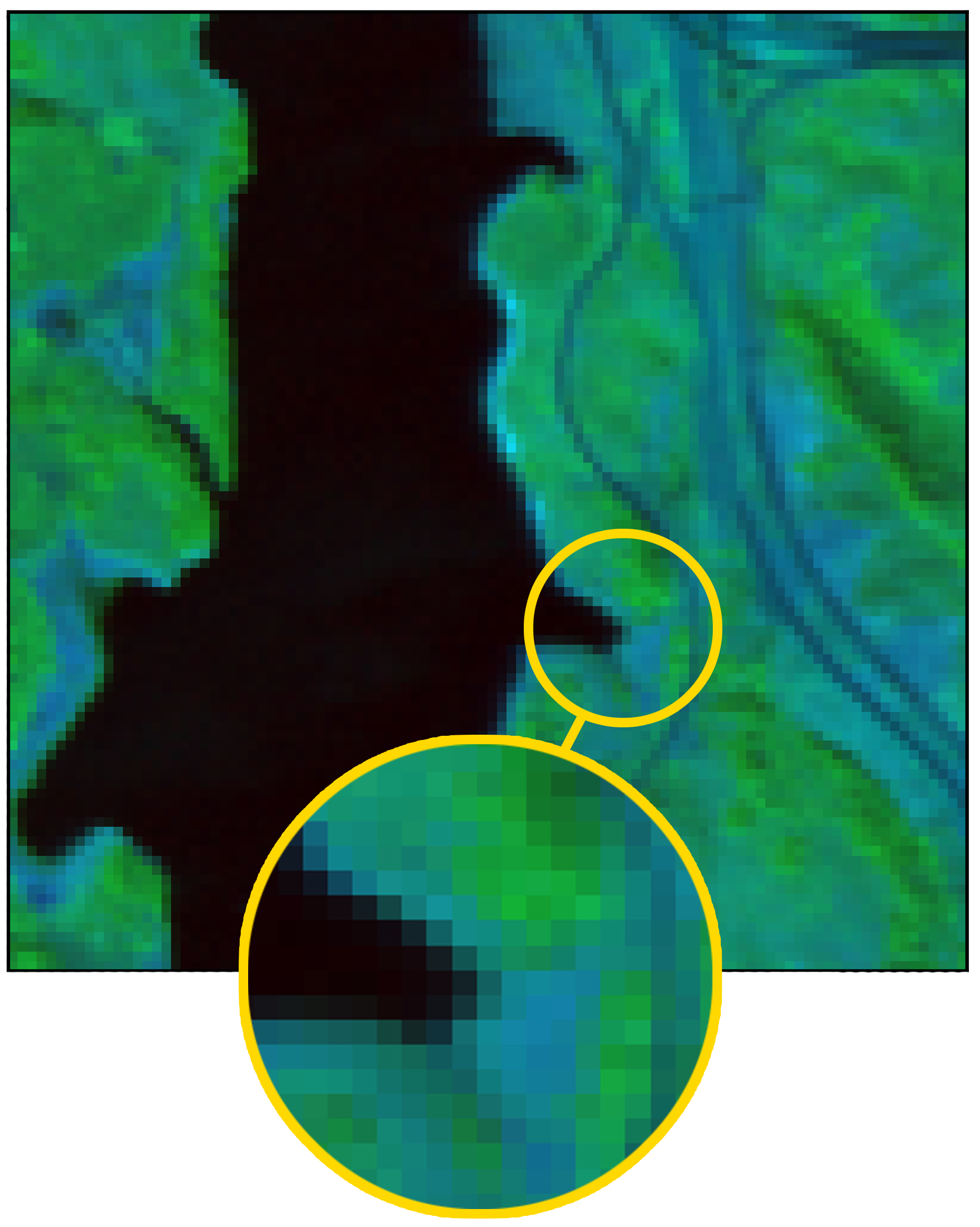}%
    \label{fig_second_case}}
    \subfloat{\includegraphics[width=0.11\textwidth]{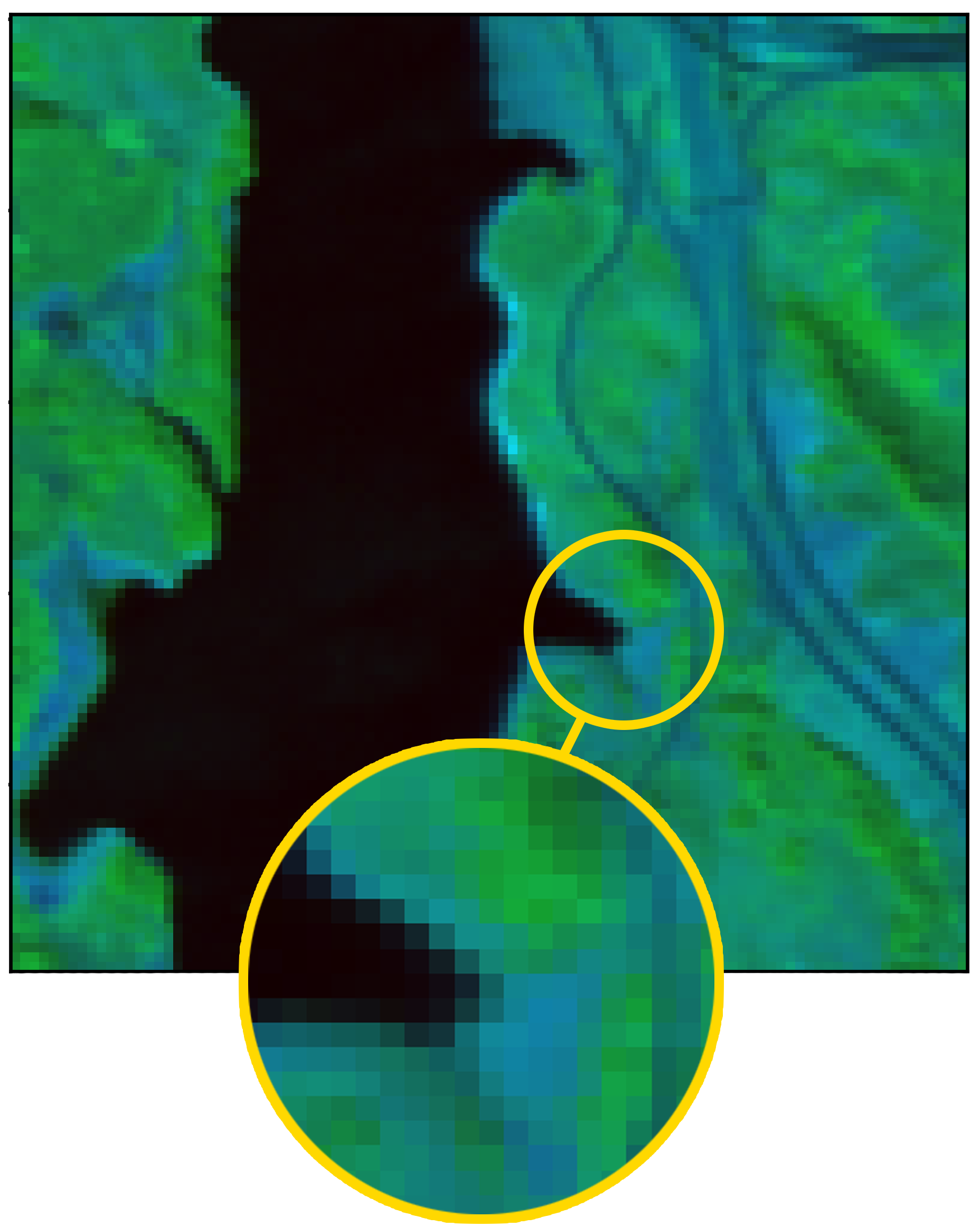}%
    \label{fig_second_case}}
    \hfil
    \subfloat{\includegraphics[width=0.11\textwidth]{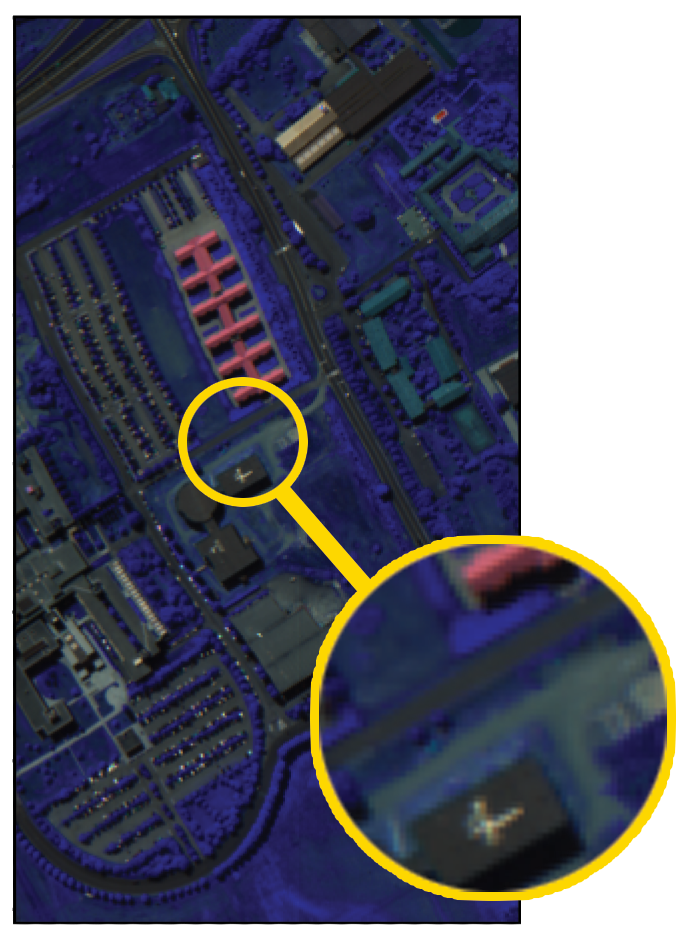}%
    \label{fig_second_case}}
    \subfloat{\includegraphics[width=0.11\textwidth]{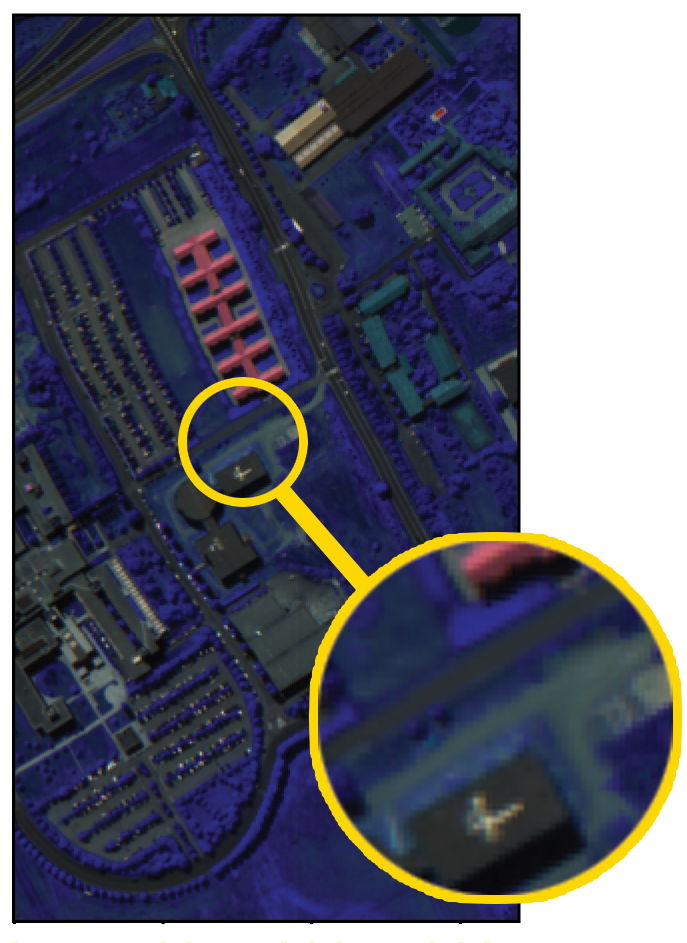}%
    \label{fig_second_case}}
    \hfil
    \subfloat{\includegraphics[width=0.11\textwidth]{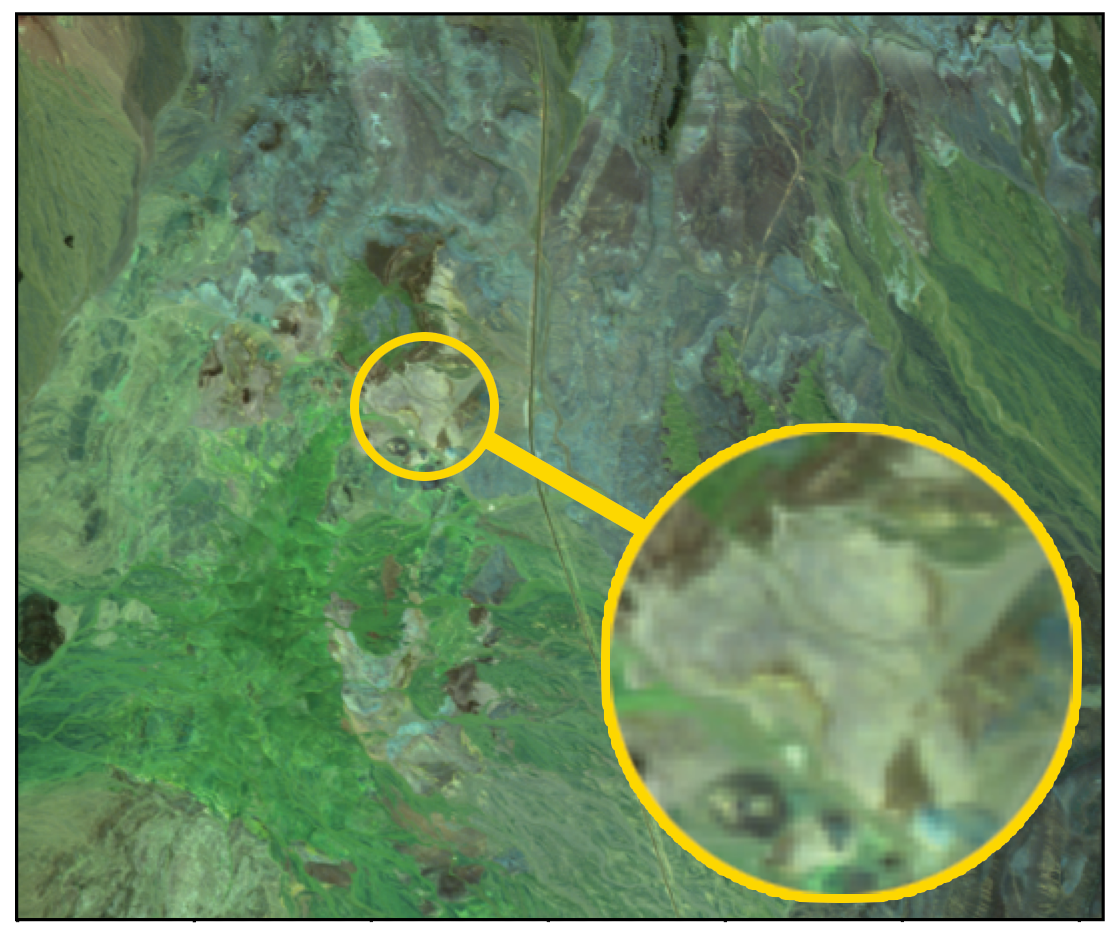}%
    \label{fig_second_case}}
    \subfloat{\includegraphics[width=0.11\textwidth]{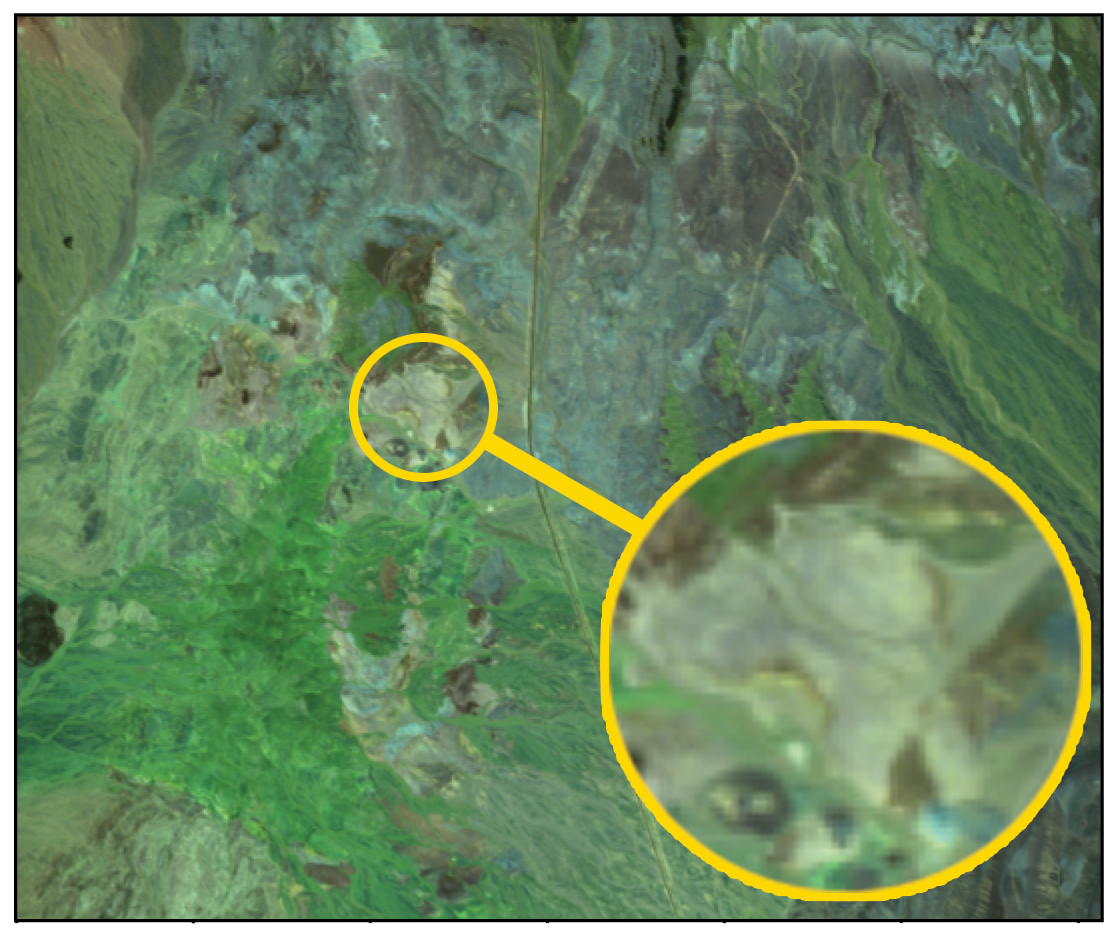}%
    \label{fig_second_case}}
    \caption{{\bf Reconstructed images} shown in pseudo-color.  (L2R) Indian Pines, Jasper Ridge, Pavia University, and Cuprite.  The image on the left in each pair is the original hyperspectral image, whereas the image on the right is the reconstructed, i.e., decompressed, hyperspectral image.  The zoomed-in portions show that the structure is preserved in the reconstructed image.  Images are shown in pseudo-color.}
    \label{fig:reconstructed-images}
\end{figure*}

\begin{table*}
\small
\centering
\begin{tabular}{|c|c|c|c|c|c|} \hline
    Dataset & Method & bppppb & compression time (Sec) & decompression time (Sec) & PSNR $\uparrow$ \\ \hline
    \multirow{4}{*}{Indian Pines} & ours & 0.1 & 243.64 & 0 & 36.98 \\
    \cline{2-6}
     & HP\_ours & 0.05 & 243.64 & 0 & 36.95 \\
    \cline{2-6}
     & ours\_sampling & 0.1 & 132.87 & 0.0005 & \textbf{39.20} \\
    \cline{2-6}
     & HP\_ours\_sampling & 0.05 & 132.87 & 0.0005 & 29.94 \\
     \cline{2-6}
     & JPEG & 0.1 & 7.353 & 3.27 & 27.47 \\
     \cline{2-6}
     & JPEG2000 & 0.1 & 0.1455 & 0.3115 & 33.58 \\
     \cline{2-6}
     & PCA-DCT & 0.1 & 1.66 & 0.04 & 32.28
     \\
    \hline
    \multirow{4}{*}{Jasper Ridge} & ours & 0.1 & 235.19 & 0.0005 & 35.77 \\
    \cline{2-6}
     & HP\_ours & 0.06 & 235.19 & 0.0005 & 35.70 \\
    \cline{2-6}
     & ours\_sampling & 0.1 & 126.33 & 0.0005 & \textbf{40.20}\\
    \cline{2-6}
     & HP\_ours\_sampling & 0.06 & 126.33 & 0.0005 & 19.58\\
    \cline{2-6}
    & JPEG & 0.1 & 3.71 & 1.62 & 24.39 \\
    \cline{2-6}
    & JPEG2000 & 0.1 & 0.138 & 0.395 & 16.75 \\
    \cline{2-6}
    & PCA-DCT & 0.1 & 1.029 & 0.027 & 25.98
    \\
    \hline
    \multirow{4}{*}{Pavia University} & ours & 0.1 & 352.74 & 0.0009 & 33.67 \\
    \cline{2-6}
     & HP\_ours & 0.05 & 352.74 & 0.0009 & 19.75 \\
    \cline{2-6}
     & ours\_sampling & 0.1 & 72.512 & 0.0004 & \textbf{38.08} \\
    \cline{2-6}
     & HP\_ours\_sampling & 0.05 & 72.512 & 0.0004 & 27.02 \\
    \cline{2-6}
     & JPEG & 0.1 & 33.86 & 14.61 & 20.86 \\
    \cline{2-6}
     & JPEG2000 & 0.1 & 0.408 & 0.628 & 17.02 \\
     \cline{2-6}
     & PCA-DCT & 0.1 & 6.525 & 0.235 & 25.121
     \\
    \hline
    \multirow{4}{*}{Cuprite} & ours & 0.06 & 1565.97 & 0.001 & 28.02 \\
    \cline{2-6}
     & HP\_ours & 0.03 & 1565.97 & 0.001 & 27.90 \\
    \cline{2-6}
     & ours\_sampling & 0.06 & 664.87 & 0.001 & \textbf{37.27} \\
    \cline{2-6}
     & HP\_ours\_sampling & 0.03 & 664.87 & 0.001 & 24.85 \\
    \cline{2-6}
    & JPEG & 0.06 & 101.195 & 45.02 & 12.88 \\
    \cline{2-6}
    & JPEG2000 & 0.06 & 1.193 & 2.476 & 15.16 \\
    \cline{2-6}
    & PCA-DCT & 0.06 & 11.67 & 0.754 & 26.75
    \\
    \hline
     \end{tabular}
 \caption{Time compression results}
\label{tbl:compression-time}
\end{table*}

\subsection{Random Sampling}
\begin{figure*}
\centering
\includegraphics[scale=0.25]{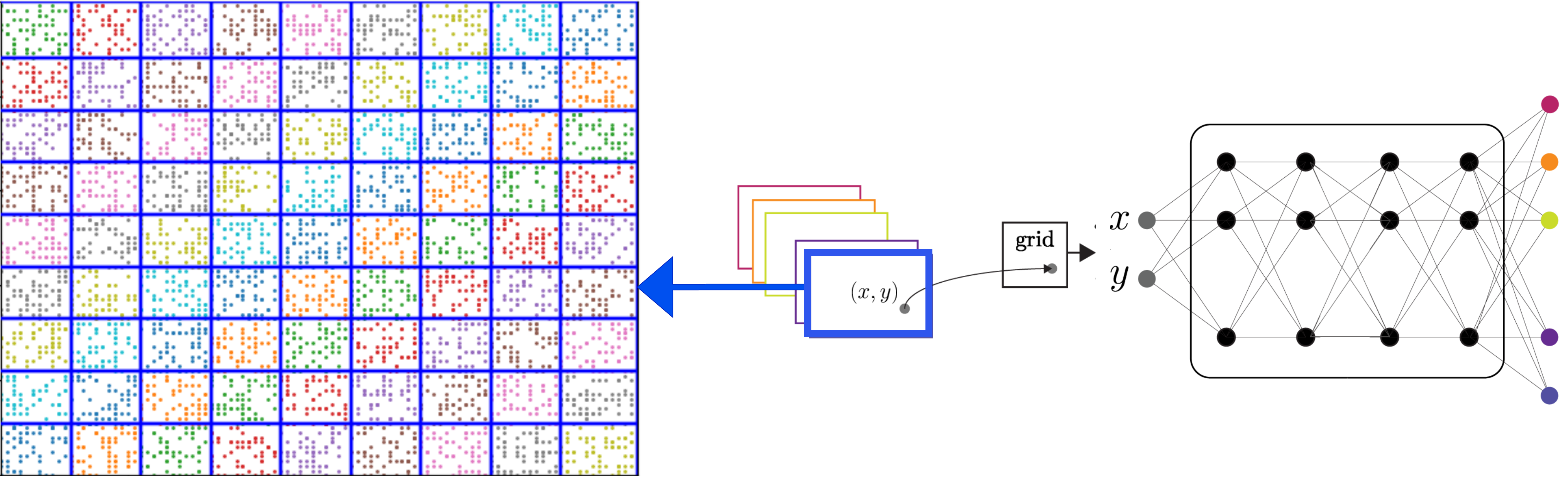}    
\caption{Compression with random sampling}
\label{fig:random_sample}
\end{figure*}

To reduce the compression time, we use the sampling method and experiment with this method on all datasets we have. Research in using sampling with implicit neural representation for hyperspectral image compression has not been done before. It is a novel work, and we will show that using sampling with implicit neural representation improves performance and speed.
Instead of feeding all the pixels as input to the neural network, we randomly pick pixels and consider those as the input to the neural network. Figure ~\ref{fig:random_sample} shows an illustration of when we want to feed the neural network with the random sampling method instead of feeding it with whole pixels. In the middle of this figure, the input image is shown with its different channels, and on the left, the selected pixels are shown with colored dots. The image is divided into several windows, then a percentage of pixels in each window will be selected. So, we have two factors here: window size and sampling rate. The window size is the size of the window where we want to collect inputs. The sampling rate is the percentage of pixels we want to collect from each window. In this example, figure ~\ref{fig:random_sample}, the window size is 9, and the sampling rate is 50 percent.
\par Table ~\ref{tbl:compression-time} shows the compression, decompression, and PSNR of our methods and JPEG, JPEG2000, and PCA-DCT for the four datasets. Here, ours\_sampling and HP\_ours\_sampling are our second methods (using sampling and INR) in full-precision and half-precision, respectively.
We fix the bpppb for each dataset, compare the compression and decompression time, and also the PSNR between our and other methods. As we can see, we improve the compression time in our methods with sampling (Ours\_sampling and HP\_ours\_sampling) and also improve the PSNR in Ours\_sampling in comparison with the methods without sampling (ours and HP\_ours).

\begin{table*}
\small
 \begin{tabular}{|c|c|c|c|c|c|c|c|c|c|c|} \hline
    Method & Dataset & Size (KB)& PSNR & bpppb & $n_h,w_h$ & Dataset & Size (KB)& PSNR & bpppb & $n_h,w_h$\\ \hline
    - & \multirow{6}{*}{Indian Pines} & 9251 & $\infty$ & 16 & -,- & \multirow{6}{*}{Jasper Ridge} & 4800 & $\infty$ & 16 & -,- \\
    \cline{1-1}
    \cline{3-6}\cline{8-11}
    JPEG &  & 115.6 & 34.085 & 0.2 & -,- & & 30 & 21.130 & 0.1 & -,- \\
    \cline{1-1}
    \cline{3-6}\cline{8-11}
    JPEG2000 &  & 115.6 & 38.098 & 0.2 & -,- &  & 30 &17.494 & 0.1 & -,-\\
    \cline{1-1}
    \cline{3-6}\cline{8-11}
    PCA-DCT &  & 115.6 & 33.173 & 0.2 & -,- &  & 30 & 26.821 & 0.1 & -,-\\
    \cline{1-1}
    \cline{3-6}\cline{8-11}
    PCA+JPEG2000 &  & 115.6 & 39.5 & 0.2 & -,- &  & 30 & - & 0.1 & -,-\\
    \cline{1-1}
    \cline{3-6}\cline{8-11}
    FPCA+JPEG2000 &  & 115.6 & 40.5 & 0.2 & -.- &  & 30 & - & 0.1 & -,-\\
    \cline{1-1}
    \cline{3-6}\cline{8-11}
    HEVC &  & 115.6 & 32 & 0.2 & -,- &  & 30 & - & 0.1 & -,-\\
    \cline{1-1}
    \cline{3-6}\cline{8-11}
    RPM &  & 115.6 & 38 & 0.2 & -,- &  & 30 & - & 0.1 & -,-\\
    \cline{1-1}
    \cline{3-6}\cline{8-11}
    3D SPECK &  & 115.6 & - & 0.2 & -,- &  & 30 & - & 0.1 & -,-\\
    \cline{1-1}
    \cline{3-6}\cline{8-11}
    3D DCT &  & 115.6 & - & 0.2 & -,- &  & 30 & - & 0.1 & -,-\\
    \cline{1-1}
    \cline{3-6}\cline{8-11}
    3D DWT+SVR &  & 115.6 & - & 0.2 & -,- &  & 30 & - & 0.1 & -,-\\
    \cline{1-1}
    \cline{3-6}\cline{8-11}
    WSRC &  & 115.6 & - & 0.2 & -,- &  & 30 & - & 0.1 & -,-\\
    \cline{1-1}
    \cline{3-6}\cline{8-11}
    ours &  & 115.6 & 40.61 & 0.2 & 15,40 &  &  30 & 35.696 & 0.1 & 10,20\\
    \cline{1-1}
    \cline{3-6}\cline{8-11}
    HP\_ours &  & 57.5 & 40.35 & 0.1 & 15,40 &  & 15 & 35.467 & 0.06 & 10,20\\
    \cline{1-1}
    \cline{3-6}\cline{8-11}
    ours\_sampling &  & 115.6 & {\bf 44.46} & 0.2 & 15,40 &  & 30 & {\bf 41.58} & 0.1 & 15,20\\
    \cline{1-1}
    \cline{3-6}\cline{8-11}
    HP\_ours\_sampling &  & 57.5 & 30.20 & 0.2 & 15,40 &  & 15 & 21.48 & 0.06 & 15,20\\
    \hline
    - & \multirow{6}{*}{Pavia University} & 42724 & $\infty$ & 16 & -,- & \multirow{6}{*}{Cuprite} & 140836 & $\infty$ & 16 & -,-\\
    \cline{1-1}
    \cline{3-6}\cline{8-11}
    JPEG &  & 267 & 20.253 & 0.1 & -,- &  & 880.2 & 24.274 & 0.1 & -,-\\
    \cline{1-1}
    \cline{3-6}\cline{8-11}
    JPEG2000 &  & 267 & 17.752 & 0.1 & -,- &  & 880.2 & 20.889 & 0.1 & -,-\\
    \cline{1-1}
    \cline{3-6}\cline{8-11}
    PCA-DCT &  & 267 & 25.436 & 0.1 & -,- &  & 880.2 & 27.302 & 0.1 & -,-\\
    \cline{1-1}
    \cline{3-6}\cline{8-11}
    PCA+JPEG2000 &  & 267 & - & 0.1 & -,- &  & 880.2 & 27.5 & 0.1 & -,-\\
    \cline{1-1}
    \cline{3-6}\cline{8-11}
    FPCA+JPEG2000 &  & 267 & - & 0.1 & -,- &  & 880.2 & - & 0.1 & -,-\\
    \cline{1-1}
    \cline{3-6}\cline{8-11}
    HEVC &  & 267 & - & 0.1 & -,- &  & 880.2 & 31 & 0.1 & -,-\\
    \cline{1-1}
    \cline{3-6}\cline{8-11}
    RPM &  & 267 & - & 0.1 & -,- &  & 880.2 & 34 & 0.1 & -,-\\
    \cline{1-1}
    \cline{3-6}\cline{8-11}
    3D SPECK &  & 267 & - & 0.1 & -,- &  & 880.2 & 27.1 & 0.1 & -,-\\
    \cline{1-1}
    \cline{3-6}\cline{8-11}
    3D DCT &  & 267 & - & 0.1 & -,- &  & 880.2 & 33.4 & 0.1 & -,-\\
    \cline{1-1}
    \cline{3-6}\cline{8-11}
    3D DWT+SVR &  & 267 & - & 0.1 & -,- &  & 880.2 & 28.20 & 0.1 & -,-\\
    \cline{1-1}
    \cline{3-6}\cline{8-11}
    WSRC &  & 267 & - & 0.1 & -,- &  & 880.2 & 35 & 0.1 & -,-\\
    \cline{1-1}
    \cline{3-6}\cline{8-11}
    ours &  & 267 & 33.749 & 0.1 & 20,60 &  & 880.2 & 28.954 & 0.1 & 25,100\\
    \cline{1-1}
    \cline{3-6}\cline{8-11}
    HP\_ours &  & 133.5 & 20.886 & 0.05 & 20,60 &  & 440.1 & 24.334 & 0.06 & 25,100\\
    \cline{1-1}
    \cline{3-6}\cline{8-11}
    ours\_sampling &  & 267 & {\bf 40.001} & 0.1 & 10,100 &  & 880.2 & {\bf 37.007} & 0.1 & 25,100\\
    \cline{1-1}
    \cline{3-6}\cline{8-11}
    HP\_ours\_sampling &  & 133.5 & 27.49 & 0.05 & 10,100 &  & 440.1 & 24.96 & 0.06 & 25,100\\ \hline
 \end{tabular}
 \caption{Compression results}
\label{tbl:compression-results}
 \end{table*}

\subsection{Compression Results}

Tables ~\ref{tbl:compression-results} lists compression results obtained by \emph{ours}, \emph{HP\_ours}, \emph{ours\_sampling}, \emph{HP\_ours\_sampling,} JPEG, JPEG2000, and PCA-DCT methods on the four datasets. We also compare our results with the learning-based methods like PCA+JPEG2000 \cite{du2007hyperspectral} and FPCA+JPEG2000 \cite{mei2018low} for the Indian Pines dataset and PCA+JPEG2000, 3D DCT \cite{yadav2018compression}, 3D DWT+SVR \cite{zikiou2020support}, and WSRC \cite{ouahioune2021enhancing} for the Cuprite dataset. Other comparisons are the comparison of our methods with HEVC \cite{sullivan2012overview} and RPM \cite{paul2016reflectance} for the Indian Pines dataset and HEVC, RPM, and 3D DCT \cite{yadav2018compression} for the cuprite dataset.
The table also shows the size of the original, uncompressed hyperspectral images.  For these results, we fix the $bpppb$ for each method, and we measure the performance of each method using PSNR. Notice that the proposed methods achieve higher PSNR values than those achieved by JPEG, JPEG2000, and PCA-DCT methods for all the datasets. The proposed methods also get better PSNR than PCA+JPEG2000, FPCA+JPEG2000, HEVC, and RPM for the Indian Pines dataset. They also achieve better PSNR than PCA+JPEG2000, HEVC, RPM, 3D SPECK, 3D DCT, 3D DWT+SVR, and WSRC for the Cuprite dataset. In Table ~\ref{tbl:compression-SSIM-cuprite-pavia}, we compare the SSIM of our methods in 0.1 bpppb with WSRC, and in 0.01 bpppb with 3D-SPECK, 3D-SPHIT \cite{fowler2007three}, 3D-WBTC \cite{bajpai20193d}, 3D-LSK \cite{ngadiran2010efficient}, 3D-NLS \cite{sudha20133d}, 3D-LMBTC \cite{bajpai2019low}, 3D-ZM-SPECK \cite{bajpai2022low} methods for Cuprite dataset. Note that our methods achieve better SSIM than other methods. We also compare the SSIM for our methods with 3D-SPHIT and 3D-DCT in 0.1 bpppb for the Pavia University dataset in this table. Our methods also achieve better SSIM than those methods.

\begin{table*}
\small
\centering
\begin{tabular}{|c|c|c|c|} \hline
    Dataset & bpppb & Method & SSIM $\uparrow$ \\ \hline
    \multirow{13}{*}{Cuprite} & \multirow{2}{*}{0.1} & WSRC & 0.75 \\
    \cline{3-4}
    &  & ours\_sampling & 0.9798 \\
    \cline{2-4}
    & \multirow{11}{*}{0.01} & 3D-SPECK & 0.142 \\ 
    \cline{3-4}
    &  & 3D-SPIHT & 0.136 \\
    \cline{3-4}
    &  & 3D-WBTC & 0.141 \\ 
    \cline{3-4}
    &  & 3D-LSK & 0.138 \\
    \cline{3-4}
    &  & 3D-NLS & 0.135 \\
    \cline{3-4}
    &  & 3D-LMBTC & 0.140 \\
    \cline{3-4}
    &  & 3D-ZM-SPECK & 0.141 \\
    \cline{3-4}
    &  & ours & 0.9565 \\
    \cline{3-4}
    &  & HP\_ours & 0.9514 \\
    \cline{3-4}
    &  & ours\_sampling & 0.9527 \\
    \cline{3-4}
    &  & HP\_ours\_sampling & 0.9390 \\
    \hline
    \multirow{6}{*}{Pavia University} & \multirow{6}{*}{0.1} & 3D-SPHIT & 0.4 \\
    \cline{3-4}
    &  & 3D-DCT & 0.85 \\
    \cline{3-4}
    &  & ours & 0.9564 \\
    \cline{3-4}
    &  & HP\_ours & 0.9527 \\
    \cline{3-4}
    &  & ours\_sampling & 0.9901 \\
    \cline{3-4}
    &  & HP\_ours\_sampling & 0.8518 \\
    \hline  
    \end{tabular}
 \caption{SSIM comparison for the Cuprite and Pavia University datasets}
\label{tbl:compression-SSIM-cuprite-pavia}
\end{table*}

\section{Conclusion}
\label{sec:conclusions}

In this work, we employ implicit neural representations to compress hyperspectral images. Multi-layer perceptron neural networks with sinusoidal activation layers are overfitted to a hyperspectral image. The network is trained to map pixel locations to pixels' spectral signatures. The parameters of the network, along with its structure, represent a compressed encoding of the original hyperspectral image. We also use a sampling method with two factors: window size and sampling rate to reduce the compression time. We have tested our approach on four datasets, and the proposed method achieves better PSNR than those achieved by JPEG, JPEG2000, and PCA-DCT methods, especially at lower bitrates. Besides, we compare our results with the learning-based methods like PCA+JPEG2000, FPCA+JPEG2000, 3D DCT, 3D DWT+SVR, and WSRC and show that we got better PSNR and SSIM than those methods. We also show that our methods with sampling achieve better speed and performance than our methods without sampling.

We plan to experiment with meta-networks to achieve smaller sizes for the compressed encodings and lower the compression times.

\bibliographystyle{IEEEtran}
\bibliography{hsi-compression}

\end{document}